\pdfoutput=1

\documentclass[11pt]{article}
\usepackage{colortbl}
\usepackage[table]{xcolor}
\usepackage[final]{acl}

\usepackage{times}
\usepackage{latexsym}

\usepackage[T1]{fontenc}

\usepackage[utf8]{inputenc}

\usepackage{microtype}

\usepackage{inconsolata}

\usepackage{setspace}

\usepackage{algorithm}
\usepackage{algorithmic}

\usepackage{graphicx}
\usepackage{times}
\usepackage{latexsym}
\usepackage[T1]{fontenc}
\usepackage[utf8]{inputenc}
\usepackage{pifont}
\newcommand{\cmark}{\ding{51}}%
\newcommand{\xmark}{\ding{55}}%

\usepackage{soul}
\usepackage{microtype}
\usepackage{tabularx}
\usepackage{xspace}
\usepackage{makecell}
\usepackage{tipa}
\usepackage{graphicx}
\usepackage{booktabs}
\usepackage{multirow}
\usepackage{caption}
\usepackage{amsmath} 
\usepackage{arydshln}
\usepackage{amssymb}  
\usepackage{xcolor}   
\usepackage{tcolorbox}
\usepackage{tabularx}
\usepackage{color}
\usepackage{bbm}
\usepackage{svg}

\title{Ranked Voting based Self-Consistency of Large Language Models}

\author{
  Weiqin Wang, Yile Wang\thanks{Corresponding author.}, Hui Huang   \\
  College of Computer Science and Software Engineering, Shenzhen University \\
  \texttt{here1swqw@gmail.com,  wangyile@szu.edu.cn, hhzhiyan@gmail.com}
}

\begin{document}
\maketitle
\begin{abstract}
\textit{Majority voting} is considered an effective method to enhance chain-of-thought reasoning, as it selects the answer with the highest ``self-consistency'' among different reasoning paths~\cite{wang2023selfconsistency}. However, previous chain-of-thought reasoning methods typically generate only a \textit{single answer} in each trial, thereby ignoring the possibility of other potential answers. As a result, these alternative answers are often overlooked in subsequent voting processes. In this work, we propose to generate \textit{ranked answers} in each reasoning process and conduct \textit{ranked voting} among multiple ranked answers from different responses, thereby making the overall self-consistency more reliable. Specifically, we use three ranked voting methods: Instant-runoff voting, Borda count voting, and mean reciprocal rank voting. We validate our methods on six datasets, including three multiple-choice and three open-ended question-answering tasks, using both advanced open-source and closed-source large language models. Extensive experimental results indicate that our proposed method outperforms the baselines, showcasing the potential of leveraging the information of ranked answers and using ranked voting to improve reasoning performance. The
code is available at \url{https://github.com/szu-tera/RankedVotingSC}.
\end{abstract}

\section{Introduction}
Large language models (LLMs) have shown strong performance in recent years~\cite{InstructGPT,gpt4,dubey2024llama,yang2024qwen2,liu2024deepseek}. Chain-of-thought prompting~\cite{cot} further improves the performance of LLMs in  commonsense~\cite{talmor-etal-2019-commonsenseqa} and mathematical ~\cite{cobbe2021training} reasoning tasks. Building on these advancements, \citet{wang2023selfconsistency} propose a majority voting based self-consistency approach, which leverages multiple reasoning paths through sampling to identify the most self-consistent answer, thereby improving the reasoning performance of LLMs.

\begin{figure}[t!]
	\centering
	\includegraphics[scale=0.96]{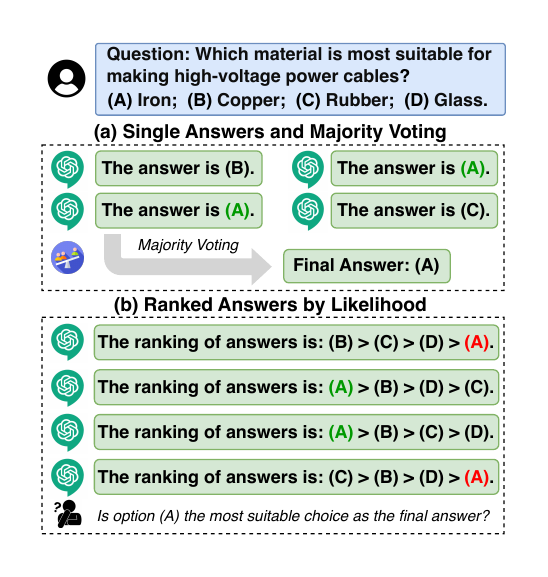}
	\caption{ Example of (a) majority voting based self-consistency among single answers~\cite{wang2023selfconsistency} and (b) ranked answers in four responses by models.}
	\label{figure:intro}
\end{figure}

An example of majority voting based self-consistency is shown in Figure~\ref{figure:intro}(a). In four responses by the model, option (A) was answered twice, (B) and (C) were each answered once, therefore (A) is considered the answer with the highest ``self-consistency''. In each response, the model replies with only one option, thereby omitting the possibility and priority of other options, which may introduce biases in the following majority voting process. In this work,  we consider obtaining ranked answers instead of only a single answer in each response and employ ranked voting based self-consistency from multiple ranked answers, which we hope can lead to more reliable self-consistency and better reasoning performance.

\begin{figure*}
  \centering
  \includegraphics[width=0.98\textwidth]{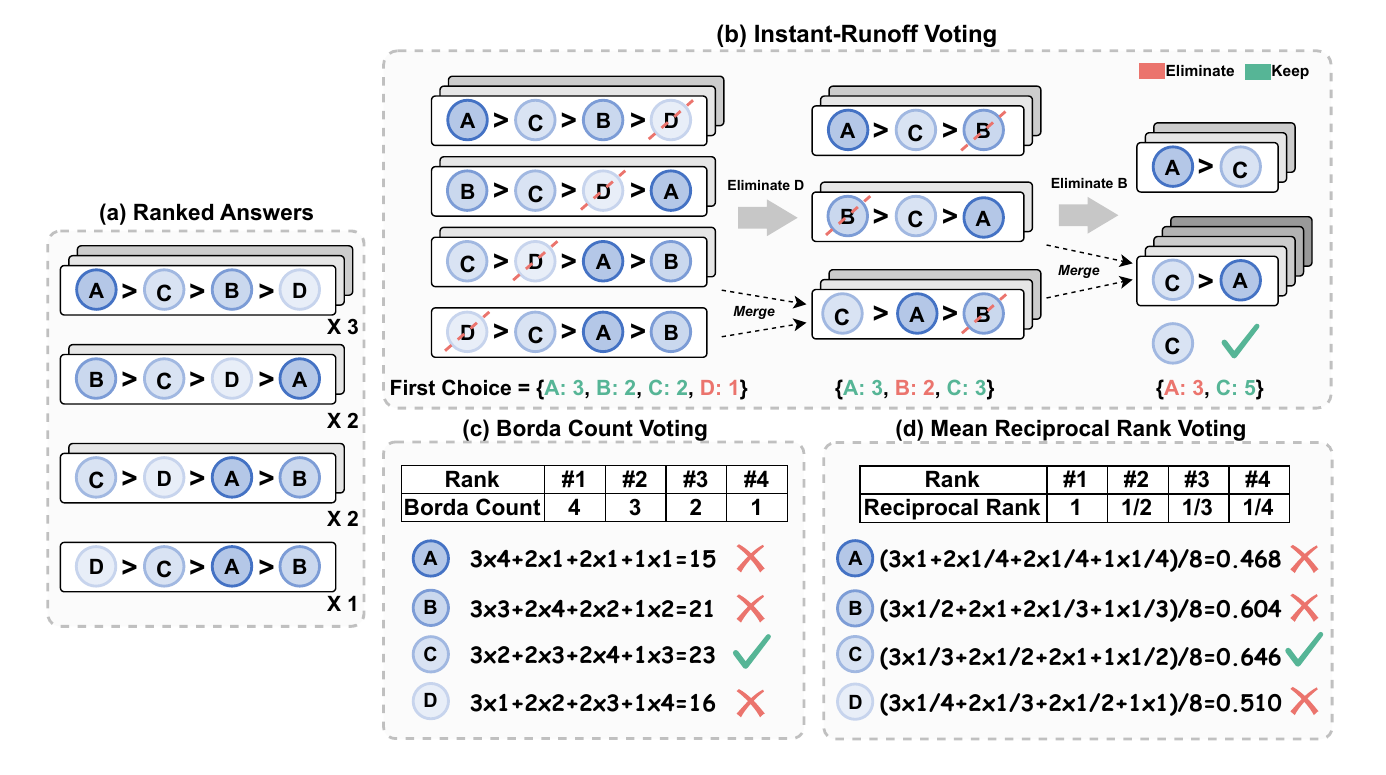}
  \caption{Examples of the procedures for three ranked voting methods. (a) The obtained ranked answers ($k=8$). (b) Instant-runoff voting. (c) Borda count voting. (d) Mean reciprocal rank voting.}
  \label{rvalg}
\end{figure*}

We show the example of ranked answers in Figure~\ref{figure:intro}(b). The model generates the possibility ranking of all options each time. In this case, we find that option (B) ranks in the top two positions in all four responses, while option (A) ranks last in two responses. Considering the overall results, option (A) may not be the best suitable final answer.

To further decide the best answer according to the ranking information, we attempt to use ranked voting to determine the most appropriate answer based on the obtained ranked answers in responses. Ranked voting has been widely used in election systems. Specifically, we use three ranking voting methods to facilitate the final answer, including instant-runoff voting~\cite{instant}, Borda count voting~\cite{borda}, and mean reciprocal rank voting. The first two methods are widely used in elections, while the latter one is related with the ranking-based MRR metric. We consider these ranked voting based approaches, compared with majority voting on single answers, can provide more reliable final answer by leveraging the ranking information of ordered answers.

We validate our ranked voting based self-consistency by using advanced LLMs, including four open-source models LLaMa-3~\cite{dubey2024llama}, Qwen-2.5~\cite{yang2024qwen2}, Gemma-2~\cite{team2024gemma}, Phi-3~\cite{abdin2024phi}, and two closed-source models GPT-3.5~\cite{gpt35} and GPT-4~\cite{gpt4}. Empirical results show that our method consistently outperforms baselines on three multiple-choice and three open-ended question-answering datasets.

\section{Method}
We first introduce the majority voting based self-consistency baseline (\S\ref{sc}) and propose obtaining the ranked answers in each LLMs' response (\S\ref{ranked-answer}). Then we provide detailed descriptions to three ranked voting methods as shown in Figure~\ref{rvalg}, including instant-runoff voting  (\S\ref{ours1}), Borda count voting  (\S\ref{ours2}), and mean reciprocal rank voting (\S\ref{ours3}). In the final we briefly present overall comparison and how to handle tie votes (\S\ref{overallcompare}).

\subsection{Majority Voting based Self-Consistency}
\label{sc}
Given a model $\mathcal{M}$ and question $\mathcal{Q}$, following chain-of-thought reasoning~\cite{cot}, we have
\begin{equation}
   \mathcal{R}, \mathcal{A} = {\mathcal{M}} (\mathcal{Q}_1,r_1,a_1...,\mathcal{Q}_n,r_n,a_n,\mathcal{Q}),
\label{eq:cot}
\end{equation}
where $\{\mathcal{Q}_j$, $r_j$, $a_j\}_{j=1}^{n}$ are questions, reasoning paths, and answers in few-shot examples, respectively. $\mathcal{R}$ and $\mathcal{A}$ are generated reasoning path and answer of the question $\mathcal{Q}$.

\citet{wang2023selfconsistency} further propose finding the most ``consistent'' answer among multiple responses. In particular, by setting a high decoding temperature (e.g., $\tau=0.7$), the final answer is the \textit{majority voting} results across $k$ possible solutions $\{\mathcal{R}_i,\mathcal{A}_i\}_{i=1}^{k}$ in totally $k$ responses:
\begin{equation}
   \mathcal{A}_\text{final}^\text{majority} = {\rm argmax}_\mathcal{A} \sum_{i=1}^{k}\mathbbm{1}(\mathcal{A}_i=\mathcal{A}).
\end{equation}

\begin{table}[t!]
    \centering
    \resizebox{\columnwidth}{!}{
    \begin{tabular}{l}
    \toprule
    \rowcolor[gray]{0.95} 
    \textbf{Multiple-Choice Question-Answering} \\
    \midrule
    \textcolor{gray}{/* Few-Shot Examples */} \\
    \makecell[l{p{8cm}}]{
    \textbf{Question:} [...] \textcolor{gray}{($\mathcal{Q}_j$)}\\
    \textbf{Answer:} [...] \textcolor{gray}{($a^{r_1}_j$$\succ$$a^{r_2}_j$$\succ$$a^{r_3}_j$$\succ$$a^{r_4}_j$)}\\
    \textcolor{gray}{/* Test Question */} \\
    \textbf{Question:} George wants to warm his hands quickly by rubbing them. Which skin surface will produce the most heat? (A) dry palms. (B) wet palms. (C) palms covered with oil. (D) palms covered with lotion. \textcolor{gray}{($\mathcal{Q}$)}\\
    \textbf{Answer:} Heat is generated by friction when two surfaces rub together. The amount of friction depends on the dryness and texture of the surfaces. Dry palms have [...]. Wet palms reduce [...].  Palms covered with oil lower [...], and palms covered with lotion reduce [...]. The ranking of options by likelihood is: \textcolor{blue}{A > B > D > C}.  \textcolor{gray}{($\mathcal{A}^{r_1}$$\succ$$\mathcal{A}^{r_2}$$\succ$$\mathcal{A}^{r_3}$$\succ$$\mathcal{A}^{r_4}$)}  \\}
    \\
    \midrule
    \midrule
    \rowcolor[gray]{0.95} 
    \textbf{Open-Ended Question-Answering} \\
    \midrule
    \textcolor{gray}{/* Few-Shot Examples */} \\
    \makecell[l{p{8cm}}]{
    \textbf{Question:} [...] \textcolor{gray}{($\mathcal{Q}_j$)}\\ 
    \textbf{Answer:} [...] \textcolor{gray}{($a^{r_1}_j$$\succ$$a^{r_2}_j$$\succ$$a^{r_3}_j$$\succ$$a^{r_4}_j$$\succ$$a^{r_5}_j$)} \\
    \textcolor{gray}{/* Test Question */} \\ 
    \textbf{Question:} The language used in Cameroon is \textcolor{gray}{($\mathcal{Q}$)}\\
    \textbf{Answer:} Cameroon is a multilingual country located in Central Africa. However, the official languages of the country are French and English. While many indigenous languages are also spoken, they do not have the same status as the official languages. The ranking of answers by likelihood is: \textcolor{blue}{French > English > Pidgin English > Fulfulde > Hausa}. \textcolor{gray}{($\mathcal{A}^{r_1}$$\succ$$\mathcal{A}^{r_2}$$\succ$$\mathcal{A}^{r_3}$$\succ$$\mathcal{A}^{r_4}$$\succ$$\mathcal{A}^{r_5}$)}}\\
    
    \bottomrule
    \end{tabular}}
    \caption{The examples to obtain ranked answers $\mathcal{A}^r$ (\textcolor{blue}{colored in blue}) on multiple-choice and open-ended question-answering tasks in few-shot settings.}
    \label{prompt-design}
\end{table}

\subsection{From Single Answer to Ranked Answers}
\label{ranked-answer}
In Eq.~\ref{eq:cot}, $\mathcal{A}$ usually indicates a \textit{single answer} such as a specific option in multiple-choice question-answering. However, it is difficult to reflect the possibility of other options from the single answer $\mathcal{A}$. Thus, we consider obtaining the \textit{ranked answers} $\mathcal{A}^r$, which contains multiple ranked candidates according to the preference of LLMs:
\begin{equation}
   \mathcal{R}, \mathcal{A}^r = {\mathcal{M}} (\mathcal{Q}_1,r_1,a_1^r...,\mathcal{Q}_n,r_n,a_n^r,\mathcal{Q}),
\label{eq:ranked-cot}
\end{equation}
where $\{\mathcal{Q}_j$, $r_j$, $a^r_j\}_{j=1}^{n}$ are questions, reasoning paths, and ranked answers in few-shot examples, respectively. $\mathcal{R}$ and $\mathcal{A}^r$ are generated reasoning path and ranked answers of the question $\mathcal{Q}$. The ranked answers $\{a^r_j\}_{j=1}^n$ and $\mathcal{A}^r$ includes $m$ ranked candidate answers:
\begin{equation}
\begin{array}{l}
\begin{aligned}
a^r_j &= a^{r_1}_j, a^{r_2}_j, ..., a^{r_m}_j,\\
\mathcal{A}^r &= \mathcal{A}^{r_1}, \mathcal{A}^{r_2}, ..., \mathcal{A}^{r_m},\\
\end{aligned}
\end{array}
\end{equation}
where $a^{r_1}_j$$\succ$$a^{r_2}_j$$\succ$...$\succ$$a^{r_m}_j$ and $\mathcal{A}^{r_1}$$\succ$$ \mathcal{A}^{r_2}$$\succ$...$\succ$$\mathcal{A}^{r_m}$ indicate that $a^{r_1}_j$ or $\mathcal{A}^{r_1}$ is the most possible answer, $a^{r_m}_j$ or $\mathcal{A}^{r_m}$ is the least possible answer in corresponding responses, respectively.

\begin{algorithm}[t!]
\caption{Instant-Runoff Voting}
\label{alg1}
\begin{algorithmic}
\STATE \textbf{Input:} Ranked answers $\textsc{Ans}=\{\mathcal{A}^r_1, ..., \mathcal{A}^r_k\}$ \\
\STATE \textbf{Output:} Final answer $\textsc{Winner}$
\WHILE{True}  
\STATE $\textsc{CurrWinner} = \text{MostFirstChoice}(\textsc{Ans})$
\IF{$\text{Count}(\textsc{CurrWinner})> k/2$}
\RETURN $\textsc{CurrWinner}$
\ENDIF
\FOR{$j$ in $1, ..., k$}
\STATE $\mathcal{A}_j^r$ = \text{EliminateTheLastOne}($\mathcal{A}_j^r$)
\ENDFOR
\STATE $\textsc{Ans}=\{\mathcal{A}^r_1, ..., \mathcal{A}^r_k\}$
\ENDWHILE
\end{algorithmic}
\end{algorithm}

\noindent \textbf{Examples of Ranked Answers.} For obtaining the ranked answers $\mathcal{A}^r$, we design the demonstrations in few-shot settings and show two examples in Table~\ref{prompt-design} on both multiple-choice and open-ended question-answering scenarios.

\noindent \textbf{Ranked Voting.} Instead of \textit{majority voting}, we leverage the information of ranked answers and get the final answer according to $k$ possible ranked solutions $\{\mathcal{A}^r_i\}_{i=1}^k = \{\mathcal{A}^{r_1}_i, \mathcal{A}^{r_2}_i, ..., \mathcal{A}^{r_m}_i\}_{i=1}^k$:
\begin{equation}
   \mathcal{A}_\text{final}^\text{ranked} = \textsc{RankVote} (\mathcal{A}^r_1, \mathcal{A}^r_2, ..., \mathcal{A}^r_k),
\end{equation}
where $\textsc{RankVote}$ indicates three ranked voting methods we used, as shown in Figure~\ref{rvalg}. We describe them in detail in the following subsections.

\subsection{Instant-Runoff Voting}
\label{ours1}
Instant-runoff voting (IRV) is a voting system that allows voters to rank candidates in order of preference~\cite{instant}. The main idea of IRV is to eliminate the candidate with the fewest votes in each round until a candidate receives a majority (more than 50\%) of the votes. In a situation where the number of votes is relatively balanced, this voting method can determine the most suitable candidate through multiple rounds of selection. We provide the IRV procedure in Algorithm~\ref{alg1} and a specific example below.

For example, consider the following ranked answers: ``$a$$\succ$$b$$\succ$$c$'' appearing on 3 responses, ``$b$$\succ$$c$$\succ$$a$'' appearing on 2 responses, and ``$c$$\succ$$a$$\succ$$b$'' appearing on 3 responses, respectively. In the initial round, candidate ``$a$'' and ``$c$'' each secure 3 first-choice votes (37.5\%), while ``$b$'' garners only 2 first-choice votes (25\%). Since no candidate achieves a majority of votes, the candidate ``$b$'' is eliminated due to the fewest first-choice votes. The second-choice votes from the responses that originally ranked ``$b$'' first are redistributed to ``$c$'', resulting in ``$c$'' accumulating a total of 5 first-choice votes (62.5\%). Meanwhile, ``$a$'' retains its 3 first-choice votes (37.5\%). Consequently, ``$c$'' emerges as the winner with a clear majority.

\subsection{Borda Count Voting}
\label{ours2}
Borda count voting (BCV) is a positional voting rule that gives each candidate a number of points (i.e., Borda count) based on their ranking~\cite{borda}. Suppose we have $m$ ranked answers $\mathcal{A}^{r_1}, \mathcal{A}^{r_2}, ..., \mathcal{A}^{r_m}$, the Borda count for candidate $\mathcal{A}$ is calculated as follows:
\begin{equation}
  \text{BordaCount}(\mathcal{A}) = \sum_{i=1}^{k} (m - \text{rank}_{\mathcal{A}}(\mathcal{A}_i^r) + 1),
\end{equation}
where $\text{rank}_{\mathcal{A}}(\mathcal{A}_i^r)$ is the rank of candidate $\mathcal{A}$ in the ranked answers $\mathcal{A}_i^r = \mathcal{A}^{r_1}_i, \mathcal{A}^{r_2}_i, ..., \mathcal{A}^{r_m}_i$ from the $i$-th response. For example, for ranked answers ``$a$$\succ$$b$$\succ$$c$'' and ``$b$$\succ$$c$$ \succ$$a$'', the Borda count for candidates ``$a$'', ``$b$'', and ``$c$'' are $3$+$1$=$4$, $2$+$3$=$5$, and $1$+$2$=$3$, respectively. The final answer $\mathcal{A}_\text{final}^\textsc{BCV}$ is the select candidate with the largest Borda count:
\begin{equation}
   \mathcal{A}_\text{final}^\text{BCV} = {\rm argmax}_\mathcal{A} \text{BordaCount}(\mathcal{A}). 
\end{equation}

\subsection{Mean Reciprocal Rank Voting}
\label{ours3}
The mean reciprocal rank (MRR) is a metric used in information retrieval to evaluate the effectiveness of search algorithms. For a sample of queries, it is calculated as the average of the multiplicative inverse of the rank of the first correct candidate: 1 for the first place, 1/2 for the second place, etc. Inspired by such rank-aware metrics, we consider evaluating the answer according to its place in multiple ranked answers, and obtain the final answer according to the MRR scores in $k$ responses, which we called mean reciprocal rank voting (MRRV). In particular, for a candidate answer $a$, we calculate the corresponding MRR score as follows:
\begin{equation}
    \text{MRR}(\mathcal{A}) = \frac{1}{k}\sum_{i=1}^{k} \frac{1}{\text{rank}_{\mathcal{A}}(\mathcal{A}_i^r)},
\end{equation}
where $\text{rank}_{\mathcal{A}}(\mathcal{A}_i^r)$ is the rank of candidate $\mathcal{A}$ in the ranked answers $\mathcal{A}_i^r = \mathcal{A}^{r_1}_i, \mathcal{A}^{r_2}_i, ..., \mathcal{A}^{r_m}_i$ from the $i$-th response. For example, for ranked answers ``$a$$\succ$$c$$\succ$$b$'' and ``$b$$\succ$$a$$\succ$$c$'' in two responses, the MRR scores for candidates ``$a$'', ``$b$'', and ``$c$'' are $\frac{1}{2}$($\frac{1}{1}$+$\frac{1}{2}$)=$0.75$, $\frac{1}{2}$($\frac{1}{3}$+$\frac{1}{1}$)=$0.66$, $\frac{1}{2}$($\frac{1}{2}$+$\frac{1}{3}$)=$0.42$, respectively. MRR score of 1.00 indicates that the corresponding candidate is always ranked first. The final answer $\mathcal{A}_\text{final}^\text{MRRV}$ is the selected candidate with the highest MRR score:
\begin{equation}
   \mathcal{A}_\text{final}^\text{MRRV} = {\rm argmax}_\mathcal{A} \text{MRR}(\mathcal{A}).
\end{equation}

\begin{table}[b!]
    \centering
    \resizebox{\columnwidth}{!}{
    \begin{tabular}{l}
    \toprule
    \rowcolor[gray]{0.95} 
    \textbf{An Example Prompt for Constructing Few-Shot} \\
    \midrule
    \makecell[l{p{8cm}}]{
    \textbf{Question:} What home entertainment equipment requires cable? (A) radio shack (B) substation (C) television (D) cabinet\\
    \textbf{Answer:} \textcolor{blue}{The most likely answer is a television, as it commonly requires cable for broadcasting or streaming services. A substation and radio shack are unrelated to home entertainment equipment, and a cabinet is purely for storage, not requiring a cable for its function.} The most likely answer is (C). The ranking of options by likelihood is: C > B > A > D.\\}
    \\    
    \bottomrule
    \end{tabular}}
    \caption{A few-shot example for question answering, showing the rationale and a candidate ranking based on semantic relevance (explanation highlighted in \textcolor{blue}{blue}).}
    \label{fewshot-construction}
\end{table}

\subsection{Construction of Few-Shot Examples}

Constructing few-shot examples for our method is straightforward. The primary criterion we follow is to ensure strong semantic relevance between the question and its relevant candidates. To demonstrate how to construction an examples, we show one example in Table~\ref{fewshot-construction}. A high-quality example should not only justify the correct answer, but also assess the plausibility of the other candidates in addressing the question, as well as their semantic relevance to it. We refer to such an example as a template. Once created, the template can be reused to construct additional examples for other questions. To scale up the few-shot example set, we begin by using existing examples as few-shot, and then leverage large language models to automatically generate additional demonstrations, requiring only minimal human verification. 

\begin{table*}[t!]
\centering
\scalebox{0.62}{
\begin{tabular}{lccccccc}
	    \toprule
        \multirow{2.5}*{\textbf{Method}}&\multicolumn{3}{c}{\textbf{Multiple-Choice QA } (Accuracy)}&\multicolumn{3}{c}{\textbf{Open-Ended QA} (Exact Match)}&\multirow{2.5}*{\textbf{Average}}\\
        \cmidrule(lr){2-4}\cmidrule(lr){5-7}
        &\textbf{CommonsenseQA}&\textbf{ARC-Challenge}&\textbf{AQUA-RAT}&\textbf{WikiData}&\textbf{Date Understanding}&\textbf{Word Unscrambling}&\\
     	\midrule
        \multicolumn{8}{c}{(\textsc{LLaMA-3.2-3B})}\\
   Few-Shot-CoT&71.99\phantom{$_{+0.00}$}&78.67\phantom{$_{+0.00}$}&58.27\phantom{$_{+0.00}$}&66.67\phantom{$_{+0.00}$}&57.72\phantom{$_{+0.00}$}&25.33\phantom{$_{+0.00}$}&59.78\phantom{$_{+0.00}$}\\
   Best-of- N&74.61\phantom{$_{+0.00}$}&80.63\phantom{$_{+0.00}$}&67.72\phantom{$_{+0.00}$}&70.83\phantom{$_{+0.00}$}&\underline{63.69}\phantom{$_{+0.00}$}&26.50\phantom{$_{+0.00}$}&64.00\phantom{$_{+0.00}$}\\
   Few-Shot-CoT-SC&73.46\phantom{$_{+0.00}$}&80.54\phantom{$_{+0.00}$}&61.81\phantom{$_{+0.00}$}&73.67\phantom{$_{+0.00}$}&60.98\phantom{$_{+0.00}$}&24.33\phantom{$_{+0.00}$}&62.47\phantom{$_{+0.00}$}\\
   Adaptive-SC&74.20\phantom{$_{+0.00}$}&80.89\phantom{$_{+0.00}$}&64.17\phantom{$_{+0.00}$}&73.33\phantom{$_{+0.00}$}&61.25\phantom{$_{+0.00}$}&26.33\phantom{$_{+0.00}$}&63.36\phantom{$_{+0.00}$}\\
   \rowcolor{gray!20}
   Instant-Runoff Voting&\textbf{74.86}$_{\textcolor{teal}{+1.40}}$&\underline{81.31}$_{\textcolor{teal}{+0.77}}$&\underline{69.29}$_{\textcolor{teal}{+7.48}}$&74.00$_{\textcolor{teal}{+0.33}}$&\textbf{64.23}$_{\textcolor{teal}{+3.25}}$&27.67$_{\textcolor{teal}{+3.34}}$&\underline{65.23}$_{\textcolor{teal}{+2.76}}$\\
   \rowcolor{gray!20}
   Borda Count Voting &\underline{74.69}$_{\textcolor{teal}{+1.23}}$&80.55$_{\textcolor{teal}{+0.01}}$&67.32$_{\textcolor{teal}{+5.51}}$&\textbf{77.00}$_{\textcolor{teal}{+3.33}}$&60.16$_{\textcolor{red}{-0.82}}$&\underline{28.33}$_{\textcolor{teal}{+4.00}}$&64.67$_{\textcolor{teal}{+2.20}}$\\
   \rowcolor{gray!20}
   Mean Reciprocal Rank Voting &74.45$_{\textcolor{teal}{+0.99}}$&\textbf{81.40}$_{\textcolor{teal}{+0.86}}$&\textbf{71.26}$_{\textcolor{teal}{+9.45}}$&\underline{76.00}$_{\textcolor{teal}{+2.32}}$&62.60$_{\textcolor{teal}{+1.62}}$&\textbf{29.00}$_{\textcolor{teal}{+4.67}}$&\textbf{65.79}$_{\textcolor{teal}{+3.32}}$\\  
    	\midrule
        \multicolumn{8}{c}{(\textsc{Qwen-2.5-3B})}\\
   Few-Shot-CoT&77.15\phantom{$_{+0.00}$}&71.42\phantom{$_{+0.00}$}&70.87\phantom{$_{+0.00}$}&69.00\phantom{$_{+0.00}$}&42.55\phantom{$_{+0.00}$}&6.67\phantom{$_{+0.00}$}&56.28\phantom{$_{+0.00}$}\\
   Best-of-N&77.97\phantom{$_{+0.00}$}&75.90\phantom{$_{+0.00}$}&75.39\phantom{$_{+0.00}$}&70.67\phantom{$_{+0.00}$}&50.14\phantom{$_{+0.00}$}&8.00\phantom{$_{+0.00}$}&59.68\phantom{$_{+0.00}$}\\
   Few-Shot-CoT-SC&77.89\phantom{$_{+0.00}$}&76.96\phantom{$_{+0.00}$}&77.95\phantom{$_{+0.00}$}&72.67\phantom{$_{+0.00}$}&47.97\phantom{$_{+0.00}$}&7.33\phantom{$_{+0.00}$}&60.13\phantom{$_{+0.00}$}\\
   Adaptive-SC&77.48\phantom{$_{+0.00}$}&78.07\phantom{$_{+0.00}$}&77.17\phantom{$_{+0.00}$}&72.67\phantom{$_{+0.00}$}&48.78\phantom{$_{+0.00}$}&7.67\phantom{$_{+0.00}$}&60.31\phantom{$_{+0.00}$}\\
   \rowcolor{gray!20}
   Instant-Runoff Voting &\textbf{78.95}$_{\textcolor{teal}{+1.06}}$&\underline{83.45}$_{\textcolor{teal}{+6.49}}$&79.13$_{\textcolor{teal}{+1.18}}$&74.67$_{\textcolor{teal}{+2.00}}$&\textbf{60.70}$_{\textcolor{teal}{+12.73}}$&\underline{11.67}$_{\textcolor{teal}{+4.34}}$&\underline{64.76}$_{\textcolor{teal}{+4.63}}$\\
   \rowcolor{gray!20}
   Borda Count Voting &78.38$_{\textcolor{teal}{+0.49}}$&83.19$_{\textcolor{teal}{+6.23}}$&\textbf{80.31}$_{\textcolor{teal}{+2.36}}$&\textbf{76.33}$_{\textcolor{teal}{+3.66}}$&58.81$_{\textcolor{teal}{+10.84}}$&11.33$_{\textcolor{teal}{+4.00}}$&64.72$_{\textcolor{teal}{+4.59}}$\\
   \rowcolor{gray!20}
   Mean Reciprocal Rank Voting&\underline{78.79}$_{\textcolor{teal}{+0.90}}$&\textbf{83.70}$_{\textcolor{teal}{+6.74}}$&\underline{79.92}$_{\textcolor{teal}{+1.97}}$&\underline{75.33}$_{\textcolor{teal}{+2.66}}$&\underline{60.43}$_{\textcolor{teal}{+12.46}}$&\textbf{12.33}$_{\textcolor{teal}{+5.00}}$&\textbf{65.08}$_{\textcolor{teal}{+4.95}}$\\  
    	\midrule
        \multicolumn{8}{c}{(\textsc{Gemma-2-2B})}\\
   Few-Shot-CoT&69.53\phantom{$_{+0.00}$}&71.33\phantom{$_{+0.00}$}&36.61\phantom{$_{+0.00}$}&72.67\phantom{$_{+0.00}$}&36.04\phantom{$_{+0.00}$}&\textbf{24.33}\phantom{$_{+0.00}$}&51.75\phantom{$_{+0.00}$}\\
   Best-of-N&70.76\phantom{$_{+0.00}$}&\underline{73.59}\phantom{$_{+0.00}$}&42.32\phantom{$_{+0.00}$}&72.17\phantom{$_{+0.00}$}&37.40\phantom{$_{+0.00}$}&20.83\phantom{$_{+0.00}$}&52.84\phantom{$_{+0.00}$}\\
   Few-Shot-CoT-SC&69.29\phantom{$_{+0.00}$}&71.42\phantom{$_{+0.00}$}&36.22\phantom{$_{+0.00}$}&73.67\phantom{$_{+0.00}$}&37.67\phantom{$_{+0.00}$}&21.67\phantom{$_{+0.00}$}&51.66\phantom{$_{+0.00}$}\\
   Adaptive-SC&70.76\phantom{$_{+0.00}$}&73.46\phantom{$_{+0.00}$}&33.46\phantom{$_{+0.00}$}&74.33\phantom{$_{+0.00}$}&37.13\phantom{$_{+0.00}$}&\underline{23.00}\phantom{$_{+0.00}$}&52.02\phantom{$_{+0.00}$}\\
   \rowcolor{gray!20}
   Instant-Runoff Voting &\textbf{71.50}$_{\textcolor{teal}{+2.21}}$&\textbf{74.74}$_{\textcolor{teal}{+3.32}}$&\textbf{44.49}$_{\textcolor{teal}{+8.27}}$&75.67$_{\textcolor{teal}{+2.00}}$&\underline{37.94}$_{\textcolor{teal}{+0.27}}$&\underline{23.00}$_{\textcolor{teal}{+1.33}}$&\textbf{54.56}$_{\textcolor{teal}{+2.90}}$\\
   \rowcolor{gray!20}
   Borda Count Voting &70.93$_{\textcolor{teal}{+1.67}}$&73.12$_{\textcolor{teal}{+1.70}}$&42.52$_{\textcolor{teal}{+6.30}}$&\textbf{77.67}$_{\textcolor{teal}{+4.00}}$&37.40$_{\textcolor{red}{-0.27}}$&22.00$_{\textcolor{teal}{+0.33}}$&53.94$_{\textcolor{teal}{+2.28}}$\\
   \rowcolor{gray!20}
   Mean Reciprocal Rank Voting&\underline{71.42}$_{\textcolor{teal}{+2.13}}$&72.18$_{\textcolor{teal}{+0.76}}$&\underline{42.91}$_{\textcolor{teal}{+6.69}}$&\underline{77.33}$_{\textcolor{teal}{+3.66}}$&\textbf{39.02}$_{\textcolor{teal}{+1.35}}$&22.67$_{\textcolor{teal}{+1.00}}$&\underline{54.26}$_{\textcolor{teal}{+2.60}}$\\  
        \midrule
        \multicolumn{8}{c}{(\textsc{Phi-3-4B})}\\
   Few-Shot-CoT&74.53\phantom{$_{+0.00}$}&86.86\phantom{$_{+0.00}$}&66.54\phantom{$_{+0.00}$}&76.00\phantom{$_{+0.00}$}&62.87\phantom{$_{+0.00}$}&\underline{24.67}\phantom{$_{+0.00}$}&65.25\phantom{$_{+0.00}$}\\
   Best-of-N&76.29\phantom{$_{+0.00}$}&86.35\phantom{$_{+0.00}$}&\underline{74.41}\phantom{$_{+0.00}$}&71.50\phantom{$_{+0.00}$}&60.70\phantom{$_{+0.00}$}&22.33\phantom{$_{+0.00}$}&65.26\phantom{$_{+0.00}$}\\
   Few-Shot-CoT-SC&75.84\phantom{$_{+0.00}$}&90.13\phantom{$_{+0.00}$}&73.62\phantom{$_{+0.00}$}&77.33\phantom{$_{+0.00}$}&\underline{66.12}\phantom{$_{+0.00}$}&23.33\phantom{$_{+0.00}$}&67.73\phantom{$_{+0.00}$}\\
   Adaptive-SC&75.76\phantom{$_{+0.00}$}&88.57\phantom{$_{+0.00}$}&71.26\phantom{$_{+0.00}$}&77.67\phantom{$_{+0.00}$}&65.31\phantom{$_{+0.00}$}&24.00\phantom{$_{+0.00}$}&67.09\phantom{$_{+0.00}$}\\
   \rowcolor{gray!20}
   Instant-Runoff Voting &78.54$_{\textcolor{teal}{+2.70}}$&\textbf{90.70}$_{\textcolor{teal}{+0.57}}$&\underline{74.41}$_{\textcolor{teal}{+0.79}}$&\underline{79.00}$_{\textcolor{teal}{+2.34}}$&\textbf{66.40}$_{\textcolor{teal}{+0.28}}$&\textbf{27.00}$_{\textcolor{teal}{+3.67}}$&\underline{69.34}$_{\textcolor{teal}{+1.61}}$\\
   \rowcolor{gray!20}
   Borda Count Voting &\underline{78.71}$_{\textcolor{teal}{+2.87}}$&88.23$_{\textcolor{red}{-1.90}}$&\textbf{75.20}$_{\textcolor{teal}{+1.58}}$&78.67$_{\textcolor{teal}{+1.34}}$&63.96$_{\textcolor{red}{-2.16}}$&\textbf{27.00}$_{\textcolor{teal}{+3.67}}$&68.63$_{\textcolor{teal}{+0.90}}$\\
   \rowcolor{gray!20}
   Mean Reciprocal Rank Voting&\textbf{78.95}$_{\textcolor{teal}{+3.11}}$&\underline{90.44}$_{\textcolor{teal}{+0.31}}$&\textbf{75.20}$_{\textcolor{teal}{+1.58}}$&\textbf{80.00}$_{\textcolor{teal}{+2.67}}$&65.58$_{\textcolor{red}{-0.54}}$&\textbf{27.00}$_{\textcolor{teal}{+3.67}}$&\textbf{69.53}$_{\textcolor{teal}{+1.80}}$\\  
     \bottomrule
\end{tabular}
}
\caption{Comparison between baselines and our method (in gray) across open-source LLMs with 2B$\sim$4B parameters. In each column, the best results are \textbf{in bold}, and the second-best results are \underline{underlined}. The subscript values represent the differences from Few-Shot-CoT-SC baseline.}
\label{table:main_results_open_models_2b_4b}
\end{table*}

\subsection{Overall Comparison and Tiebreaker Rule}
\label{overallcompare}
\noindent\textbf{Comparison of Methods.} The ranked voting methods IRV, BCV, and MRRV are different from each other. Specifically, IRV can be regarded as an elimination-based ranked voting, while BCV and MRRV can be seen as ranked voting with different weighting schemes. In Figure~\ref{rvalg}, all three methods arrive at the same final answer (C). However, in practice, their results may vary due to the specific outcomes of ranked answers.

\noindent\textbf{Tiebreaker Rules.} Although ties are rare in voting outcomes, they may occur particularly for complex questions. Compared with self-consistency baseline, our ranking-based approach mitigates this issue  by incorporating preference ranking information (see details in $\S$\ref{section:analysis}). To ensure rigor, we handle tie scenarios through confidence scoring for all methods, including the baseline: when multiple candidates receive equal votes, we compute each answer's confidence score $\mathcal{S}_i = \sum_{t=1}^{n} \log(p(\mathcal{C}_{i,t}))$, where $\mathcal{C}_{i,t}$ denotes the $t$-th token's predicted probability in the $i$-th candidate. The final selection chooses the candidate with the highest sum of logarithmic probabilities. For closed-source models, we defer to the model itself to select the best answer from the candidates.

\section{Experiments}

\subsection{Settings}

\noindent\textbf{Datasets.} We validate our method on six tasks across different domains. The three multiple-choice QA tasks include: \textbf{CommonsenseQA}~\cite{talmor-etal-2019-commonsenseqa} is a multiple-choice question answering dataset that requires different types of commonsense knowledge. \textbf{ARC-Challenge}~\cite{clark2018think} includes genuine grade-school level, multiple-choice science questions. \textbf{AQUA-RAT}~\cite{ling-etal-2017-program} is a multiple-choice QA dataset which involves five-options algebraic word problems with rationales. We also evaluate on three open-ended QA tasks form Big-bench~\cite{srivastava2023beyond}: \textbf{WikiData} involves performing open-domain cloze-style question answering. \textbf{Date Understanding} aims to measure models' ability to understand date-related information. \textbf{Word Unscrambling} asks models to unscramble the given letters to form an English word. We list examples of each dataset in Appendix~\ref{appendix:examples}.

\begin{table*}[t!]
  \centering
  \scalebox{0.62}{
  \begin{tabular}{lccccccc}
        \toprule
          \multirow{2.5}*{\textbf{Method}}&\multicolumn{3}{c}{\textbf{Multiple-Choice QA } (Accuracy)}&\multicolumn{3}{c}{\textbf{Open-Ended QA} (Exact Match)}&\multirow{2.5}*{\textbf{Average}}\\
          \cmidrule(lr){2-4}\cmidrule(lr){5-7}
          &\textbf{CommonsenseQA}&\textbf{ARC-Challenge}&\textbf{AQUA-RAT}&\textbf{WikiData}&\textbf{Date Understanding}&\textbf{Word Unscrambling}&\\
        \midrule
        \multicolumn{8}{c}{(\textsc{LLaMA-3-8B})}\\
  Few-Shot-CoT&78.13\phantom{$_{+0.00}$}&84.13\phantom{$_{+0.00}$}&62.99\phantom{$_{+0.00}$}&78.33\phantom{$_{+0.00}$}&70.19\phantom{$_{+0.00}$}&25.67\phantom{$_{+0.00}$}&66.67\phantom{$_{+0.00}$}\\
  Best-of-N&79.03\phantom{$_{+0.00}$}&85.15\phantom{$_{+0.00}$}&72.44\phantom{$_{+0.00}$}&77.83\phantom{$_{+0.00}$}&69.92\phantom{$_{+0.00}$}&26.83\phantom{$_{+0.00}$}&68.53\phantom{$_{+0.00}$}\\
   Few-Shot-CoT-SC&78.71\phantom{$_{+0.00}$}&86.77\phantom{$_{+0.00}$}&66.93\phantom{$_{+0.00}$}&80.00\phantom{$_{+0.00}$}&\underline{71.82}\phantom{$_{+0.00}$}&24.00\phantom{$_{+0.00}$}&68.04\phantom{$_{+0.00}$}\\
   Adaptive-SC&78.95\phantom{$_{+0.00}$}&86.95\phantom{$_{+0.00}$}&68.50\phantom{$_{+0.00}$}&79.67\phantom{$_{+0.00}$}&71.00\phantom{$_{+0.00}$}&25.33\phantom{$_{+0.00}$}&68.40\phantom{$_{+0.00}$}\\
   \rowcolor{gray!20}
   Instant-Runoff Voting &79.12$_{\textcolor{teal}{+0.41}}$&\underline{87.29}$_{\textcolor{teal}{+0.52}}$&\underline{74.02}$_{\textcolor{teal}{+7.09}}$&80.00$_{\textcolor{teal} {+0.00}}$&\textbf{72.09}$_{\textcolor{teal}{+0.27}}$&\underline{35.00}$_{\textcolor{teal}{+11.00}}$&\underline{71.25}$_{\textcolor{teal}{+3.21}}$\\
   \rowcolor{gray!20}
   Borda Count Voting &\textbf{79.69}$_{\textcolor{teal}{+0.98}}$&87.12$_{\textcolor{teal}{+0.35}}$&70.87$_{\textcolor{teal}{+3.94}}$&\underline{80.33}$_{\textcolor{teal}{+0.33}}$&71.00$_{\textcolor{red}{-0.82}}$&\textbf{35.33}$_{\textcolor{teal}{+11.33}}$&70.72$_{\textcolor{teal}{+2.68}}$\\
   \rowcolor{gray!20}
   Mean Reciprocal Rank Voting&\underline{79.36}$_{\textcolor{teal}{+0.65}}$&\textbf{87.46}$_{\textcolor{teal}{+0.69}}$&\textbf{75.20}$_{\textcolor{teal}{+8.27}}$&\textbf{80.67}$_{\textcolor{teal}{+0.67}}$&71.27$_{\textcolor{red}{-0.45}}$&\textbf{35.33}$_{\textcolor{teal}{+11.33}}$&\textbf{71.55}$_{\textcolor{teal}{+3.51}}$\\  
         \midrule
         \multicolumn{8}{c}{(\textsc{Qwen-2.5-7B})}\\
  Few-Shot-CoT&83.29\phantom{$_{+0.00}$}&87.80\phantom{$_{+0.00}$}&80.31\phantom{$_{+0.00}$}&68.00\phantom{$_{+0.00}$}&68.29\phantom{$_{+0.00}$}&17.00\phantom{$_{+0.00}$}&67.45\phantom{$_{+0.00}$}\\
  Best-of-N&84.32\phantom{$_{+0.00}$}&89.04\phantom{$_{+0.00}$}&82.87\phantom{$_{+0.00}$}&69.00\phantom{$_{+0.00}$}&70.05\phantom{$_{+0.00}$}&23.17\phantom{$_{+0.00}$}&69.74\phantom{$_{+0.00}$}\\
  Few-Shot-CoT-SC&84.28\phantom{$_{+0.00}$}&\underline{89.16}\phantom{$_{+0.00}$}&84.65\phantom{$_{+0.00}$}&70.33\phantom{$_{+0.00}$}&\underline{70.73}\phantom{$_{+0.00}$}&21.00\phantom{$_{+0.00}$}&70.03\phantom{$_{+0.00}$}\\
   Adaptive-SC&84.68\phantom{$_{+0.00}$}&\textbf{89.25}\phantom{$_{+0.00}$}&83.46\phantom{$_{+0.00}$}&71.33\phantom{$_{+0.00}$}&70.46\phantom{$_{+0.00}$}&22.33\phantom{$_{+0.00}$}&70.25\phantom{$_{+0.00}$}\\
   \rowcolor{gray!20}
   Instant-Runoff Voting &\underline{84.77}$_{\textcolor{teal}{+0.49}}$&\underline{89.16}$_{\textcolor{teal}{+0.00}}$&\underline{85.04}$_{\textcolor{teal}{+0.39}}$&77.67$_{\textcolor{teal}{+7.34}}$&\textbf{71.27}$_{\textcolor{teal}{+0.54}}$&\textbf{24.00}$_{\textcolor{teal}{+3.00}}$&\underline{71.98}$_{\textcolor{teal}{+1.95}}$\\
   \rowcolor{gray!20}
   Borda Count Voting &84.52$_{\textcolor{teal}{+0.24}}$&\textbf{89.25}$_{\textcolor{teal}{+0.09}}$&84.25$_{\textcolor{red}{-0.40}}$&\textbf{79.33}$_{\textcolor{teal}{+9.00}}$&\underline{70.73}$_{\textcolor{teal}{+0.00}}$&\underline{23.33}$_{\textcolor{teal}{+2.33}}$&71.90$_{\textcolor{teal}{+1.87}}$\\
   \rowcolor{gray!20}
   Mean Reciprocal Rank Voting&\textbf{84.93}$_{\textcolor{teal}{+0.65}}$&87.80$_{\textcolor{red}{-1.36}}$&\textbf{85.83}$_{\textcolor{teal}{+1.18}}$&\underline{78.67}$_{\textcolor{teal}{+8.34}}$&\textbf{71.27}$_{\textcolor{teal}{+0.54}}$&\textbf{24.00}$_{\textcolor{teal}{+3.00}}$&\textbf{72.08}$_{\textcolor{teal}{+2.05}}$\\  
         \midrule
         \multicolumn{8}{c}{(\textsc{Gemma-2-9B})}\\
  Few-Shot-CoT&80.75\phantom{$_{+0.00}$}&88.05\phantom{$_{+0.00}$}&63.39\phantom{$_{+0.00}$}&78.00\phantom{$_{+0.00}$}&77.24\phantom{$_{+0.00}$}&45.67\phantom{$_{+0.00}$}&72.18\phantom{$_{+0.00}$}\\
  Best-of-N&80.22\phantom{$_{+0.00}$}&88.52\phantom{$_{+0.00}$}&71.46\phantom{$_{+0.00}$}&77.50\phantom{$_{+0.00}$}&76.29\phantom{$_{+0.00}$}&43.17\phantom{$_{+0.00}$}&72.86\phantom{$_{+0.00}$}\\
  Few-Shot-CoT-SC&80.10\phantom{$_{+0.00}$}&\underline{89.59}\phantom{$_{+0.00}$}&67.32\phantom{$_{+0.00}$}&81.33\phantom{$_{+0.00}$}&78.05\phantom{$_{+0.00}$}&44.33\phantom{$_{+0.00}$}&73.45\phantom{$_{+0.00}$}\\
   Adaptive-SC&81.41\phantom{$_{+0.00}$}&86.77\phantom{$_{+0.00}$}&66.93\phantom{$_{+0.00}$}&80.67\phantom{$_{+0.00}$}&77.78\phantom{$_{+0.00}$}&45.67\phantom{$_{+0.00}$}&73.20\phantom{$_{+0.00}$}\\
   \rowcolor{gray!20}
   Instant-Runoff Voting &\textbf{83.21}$_{\textcolor{teal}{+3.11}}$&\textbf{89.68}$_{\textcolor{teal}{+0.09}}$&\textbf{72.83}$_{\textcolor{teal}{+5.51}}$&82.67$_{\textcolor{teal}{+1.34}}$&\underline{78.59}$_{\textcolor{teal}{+0.54}}$&\textbf{46.33}$_{\textcolor{teal}{+2.00}}$&\textbf{75.55}$_{\textcolor{teal}{+2.10}}$\\
   \rowcolor{gray!20}
   Borda Count Voting &82.96$_{\textcolor{teal}{+1.86}}$&89.33$_{\textcolor{red}{-0.26}}$&70.87$_{\textcolor{teal}{+3.55}}$&\textbf{83.33}$_{\textcolor{teal}{+2.00}}$&\textbf{79.40}$_{\textcolor{teal}{+1.35}}$&\underline{46.00}$_{\textcolor{teal}{+1.67}}$&75.31$_{\textcolor{teal}{+1.86}}$\\
   \rowcolor{gray!20}
   Mean Reciprocal Rank Voting&\underline{83.05}$_{\textcolor{teal}{+2.95}}$&89.25$_{\textcolor{red}{-0.34}}$&72.44$_{\textcolor{teal}{+5.12}}$&\underline{83.00}$_{\textcolor{teal}{+1.67}}$&78.32$_{\textcolor{teal}{+0.27}}$&\textbf{46.33}$_{\textcolor{teal}{+2.00}}$&\underline{75.40}$_{\textcolor{teal}{+1.95}}$\\  
         \midrule
         \multicolumn{8}{c}{(\textsc{Phi-3-7B})}\\
    Few-Shot-CoT&80.43\phantom{$_{+0.00}$}&91.47\phantom{$_{+0.00}$}&65.75\phantom{$_{+0.00}$}&79.33\phantom{$_{+0.00}$}&68.56\phantom{$_{+0.00}$}&27.67\phantom{$_{+0.00}$}&68.87\phantom{$_{+0.00}$}\\
    Best-of-N&80.79\phantom{$_{+0.00}$}&91.30\phantom{$_{+0.00}$}&\textbf{77.17}\phantom{$_{+0.00}$}&77.17\phantom{$_{+0.00}$}&73.04\phantom{$_{+0.00}$}&\underline{29.67}\phantom{$_{+0.00}$}&71.52\phantom{$_{+0.00}$}\\
    Few-Shot-CoT-SC&81.24\phantom{$_{+0.00}$}&91.89\phantom{$_{+0.00}$}&73.62\phantom{$_{+0.00}$}&79.33\phantom{$_{+0.00}$}&73.71\phantom{$_{+0.00}$}&28.33\phantom{$_{+0.00}$}&71.35\phantom{$_{+0.00}$}\\
   Adaptive-SC&81.16\phantom{$_{+0.00}$}&\underline{91.98}\phantom{$_{+0.00}$}&74.02\phantom{$_{+0.00}$}&79.00\phantom{$_{+0.00}$}&73.17\phantom{$_{+0.00}$}&\underline{29.67}\phantom{$_{+0.00}$}&71.50\phantom{$_{+0.00}$}\\
   \rowcolor{gray!20}
   Instant-Runoff Voting &81.98$_{\textcolor{teal}{+0.74}}$&\textbf{92.24}$_{\textcolor{teal}{+0.35}}$&\textbf{77.17}$_{\textcolor{teal}{+3.55}}$&\underline{80.00}$_{\textcolor{teal}{+0.67}}$&\underline{75.34}$_{\textcolor{teal}{+1.63}}$&\textbf{30.33}$_{\textcolor{teal}{+2.00}}$&\textbf{72.84}$_{\textcolor{teal}{+1.49}}$\\
   \rowcolor{gray!20}
   Borda Count Voting &\textbf{82.31}$_{\textcolor{teal}{+1.07}}$&91.81$_{\textcolor{red}{-0.08}}$&\underline{75.20}$_{\textcolor{teal}{+1.58}}$&\underline{80.00}$_{\textcolor{teal}{+0.67}}$&\textbf{75.61}$_{\textcolor{teal}{+1.90}}$&28.33$_{\textcolor{teal}{+0.00}}$&72.21$_{\textcolor{teal}{+0.86}}$\\
   \rowcolor{gray!20}
   Mean Reciprocal Rank Voting&\underline{82.15}$_{\textcolor{teal}{+0.91}}$&\underline{91.98}$_{\textcolor{teal}{+0.09}}$&\textbf{77.17}$_{\textcolor{teal}{+3.55}}$&\textbf{81.00}$_{\textcolor{teal}{+1.67}}$&\underline{75.34}$_{\textcolor{teal}{+1.63}}$&28.67$_{\textcolor{teal}{+0.34}}$&\underline{72.72}$_{\textcolor{teal}{+1.37}}$\\  
       \bottomrule
  \end{tabular}
  }
  \caption{Comparison between baselines and our method (in gray) across open-source LLMs with 7B$\sim$9B parameters.}
  \label{table:main_results_open_models_7b_9b}
\end{table*}

\vspace{0.7mm}
\noindent\textbf{Models.} We compare with widely-used LLMs, including open-source models LLaMA-3~\cite{dubey2024llama}, Qwen-2.5~\cite{yang2024qwen2}, Gemma-2~\cite{team2024gemma}, and Phi-3~\cite{abdin2024phi}, as well as closed-source GPT-3.5-turbo~\cite{gpt35} and GPT-4-turbo~\cite{gpt4} models. For open-source models, we use both lightweight (2B$\sim$4B) and medium-sized (7B$\sim$9B) models for comprehensive evaluations, and the checkpoints are listed in Appendix~\ref{appendix:checkpoints}.

\vspace{0.7mm}
\noindent\textbf{Baseline Methods.} We compare with the following baselines: \textbf{Few-Shot-CoT}~\cite{cot} with chain-of-thought reasoning. \textbf{Best-of-N}~\cite{best-of-n} involves sampling multiple solutions and selecting the best one based on the scores of a fine-tuned Qwen2.5-7B~\cite{yang2024qwen2} reward model. \textbf{Few-Shot-CoT-SC}~\cite{wang2023selfconsistency} uses majority voting for aggregating multiple answers. \textbf{Adaptive-SC}~\cite{aggarwal-etal-2023-lets} uses dynamic number of responses instead of fixed ones.

\vspace{0.7mm}
\noindent\textbf{Implementation Details.} For all tasks, we use LM-evaluation-harness~\cite{eval-harness} for fair comparison. Following~\citet{cot}, we set the number of few-shot examples $n=8$ and decoding temperature $\tau=0.7$ for baselines and our methods. Considering the budget constraints, we set the number of responses $k=8$ in main experiments.

\subsection{Main Results}

\noindent\textbf{2B$\sim$4B Open-Source LLMs.} The results are shown in Table~\ref{table:main_results_open_models_2b_4b}. Overall, our method demonstrates improved average performance across various datasets and models, surpassing all baseline methods. Compared to the Few-Shot-CoT-SC, our three ranked voting methods achieve an average improvement of 3.32\% with LLaMA-3.2-3B and 4.95\% with Qwen-2.5-3B. Our method also outperforms Best-of-N, which relies on a reward model to enhance generation quality and incurs additional computational overhead.

Across different tasks, we achieve the largest improvement when performing data understanding tasks using Qwen-2.5-3B. Compared to Few-Shot-CoT-SC, the improvement can reach 10.84\% to 12.46\%. This suggests that, in contrast to providing a single answer for majority voting, offering multiple candidate answers along with their ranking and conducting ranked voting can offer more useful context. 

\begin{table*}[t!]
    \centering
    \scalebox{0.62}{
    \begin{tabular}{lccccccc}
            \toprule
            \multirow{2.5}*{\textbf{Method}}&\multicolumn{3}{c}{\textbf{Multiple-Choice QA } (Accuracy)}&\multicolumn{3}{c}{\textbf{Open-Ended QA} (Exact Match)}&\multirow{2.5}*{\textbf{Average}}\\
            \cmidrule(lr){2-4}\cmidrule(lr){5-7}
            &\textbf{CommonsenseQA}&\textbf{ARC-Challenge}&\textbf{AQUA-RAT}&\textbf{WikiData}&\textbf{Date Understanding}&\textbf{Word Unscrambling}&\\
            \midrule
            \multicolumn{8}{c}{(\textsc{gpt-3.5-turbo-0125})}\\
       Few-Shot-CoT&79.85\phantom{$_{+0.00}$}&86.65\phantom{$_{+0.00}$}&58.27\phantom{$_{+0.00}$}&78.00\phantom{$_{+0.00}$}&58.27\phantom{$_{+0.00}$}&49.50\phantom{$_{+0.00}$}&68.42\phantom{$_{+0.00}$}\\
       Best-of-N&80.34\phantom{$_{+0.00}$}&87.86\phantom{$_{+0.00}$}&\textbf{72.44}\phantom{$_{+0.00}$}&79.83\phantom{$_{+0.00}$}&61.11\phantom{$_{+0.00}$}&58.17\phantom{$_{+0.00}$}&73.29\phantom{$_{+0.00}$}\\
       Few-Shot-CoT-SC&79.61\phantom{$_{+0.00}$}&87.54\phantom{$_{+0.00}$}&69.49\phantom{$_{+0.00}$}&79.00\phantom{$_{+0.00}$}&57.99\phantom{$_{+0.00}$}&54.50\phantom{$_{+0.00}$}&71.36\phantom{$_{+0.00}$}\\   
       Adaptive-SC&79.77\phantom{$_{+0.00}$}&88.25\phantom{$_{+0.00}$}&70.67\phantom{$_{+0.00}$}&79.33\phantom{$_{+0.00}$}&58.40\phantom{$_{+0.00}$}&54.50\phantom{$_{+0.00}$}&71.82\phantom{$_{+0.00}$}\\
       \rowcolor{gray!20}
       Instant-Runoff Voting &80.84$_{\textcolor{teal}{+1.23}}$&\underline{89.33}$_{\textcolor{teal}{+1.79}}$&70.67$_{\textcolor{teal}{+1.18}}$&\underline{82.00}$_{\textcolor{teal}{+3.00}}$&\textbf{68.97}$_{\textcolor{teal}{+10.98}}$&\underline{66.50}$_{\textcolor{teal}{+12.00}}$&\underline{76.30}$_{\textcolor{teal}{+4.94}}$\\
       \rowcolor{gray!20}
       Borda Count Voting &\underline{80.92}$_{\textcolor{teal}{+1.31}}$&\textbf{89.59}$_{\textcolor{teal}{+2.05}}$&68.90$_{\textcolor{red}{-0.59}}$&81.50$_{\textcolor{teal}{+2.50}}$&68.70$_{\textcolor{teal}{+10.71}}$&\textbf{67.67}$_{\textcolor{teal}{+13.17}}$&76.21$_{\textcolor{teal}{+4.85}}$\\
       \rowcolor{gray!20}
       Mean Reciprocal Rank Voting&\textbf{81.00}$_{\textcolor{teal}{+1.39}}$&89.25$_{\textcolor{teal}{+1.71}}$&\underline{71.06}$_{\textcolor{teal}{+1.57}}$&\textbf{82.33}$_{\textcolor{teal}{+3.33}}$&\underline{68.83}$_{\textcolor{teal}{+10.84}}$&\textbf{67.67}$_{\textcolor{teal}{+13.17}}$&\textbf{76.69}$_{\textcolor{teal}{+5.33}}$\\  
             \midrule
            \multicolumn{8}{c}{(\textsc{gpt-4-turbo-2024-04-09})}\\
       Few-Shot-CoT&86.32\phantom{$_{+0.00}$}&93.69\phantom{$_{+0.00}$}&80.31\phantom{$_{+0.00}$}&82.00\phantom{$_{+0.00}$}&88.08\phantom{$_{+0.00}$}&78.67\phantom{$_{+0.00}$}&84.84\phantom{$_{+0.00}$}\\
      Best-of-N&\underline{87.39}\phantom{$_{+0.00}$}&94.70\phantom{$_{+0.00}$}&85.43\phantom{$_{+0.00}$}&82.33\phantom{$_{+0.00}$}&86.99\phantom{$_{+0.00}$}&\underline{81.67}\phantom{$_{+0.00}$}&86.42\phantom{$_{+0.00}$}\\
       Few-Shot-CoT-SC&\underline{87.39}\phantom{$_{+0.00}$}&95.30\phantom{$_{+0.00}$}&\underline{86.22}\phantom{$_{+0.00}$}&\underline{83.00}\phantom{$_{+0.00}$}&\underline{88.35}\phantom{$_{+0.00}$}&81.33\phantom{$_{+0.00}$}&86.93\phantom{$_{+0.00}$}\\
       Adaptive-SC&87.22\phantom{$_{+0.00}$}&95.21\phantom{$_{+0.00}$}&\underline{86.22}\phantom{$_{+0.00}$}&82.67\phantom{$_{+0.00}$}&\textbf{88.62}\phantom{$_{+0.00}$}&81.00\phantom{$_{+0.00}$}&86.82\phantom{$_{+0.00}$}\\
       \rowcolor{gray!20}
       Instant-Runoff Voting &\textbf{87.47}$_{\textcolor{teal}{+0.08}}$&\textbf{97.01}$_{\textcolor{teal}{+1.71}}$&\textbf{86.61}$_{\textcolor{teal}{+0.39}}$&\underline{83.00}$_{\textcolor{teal}{+0.00}}$&\underline{88.35}$_{\textcolor{teal}{+0.00}}$&\textbf{82.00}$_{\textcolor{teal}{+0.67}}$&\textbf{87.41}$_{\textcolor{teal}{+0.48}}$\\
       \rowcolor{gray!20}
       Borda Count Voting &87.31$_{\textcolor{red}{-0.08}}$&\underline{96.84}$_{\textcolor{teal}{+1.54}}$&85.04$_{\textcolor{red}{-1.18}}$&\textbf{83.33}$_{\textcolor{teal}{+0.33}}$&\underline{88.35}$_{\textcolor{teal}{+0.00}}$&\underline{81.67}$_{\textcolor{teal}{+0.34}}$&87.09$_{\textcolor{teal}{+0.16}}$\\
       \rowcolor{gray!20}
       Mean Reciprocal Rank Voting&87.31$_{\textcolor{red}{-0.08}}$&\textbf{97.01}$_{\textcolor{teal}{+1.71}}$&\underline{86.22}$_{\textcolor{teal}{+0.00}}$&\underline{83.00}$_{\textcolor{teal}{+0.00}}$&\textbf{88.62}$_{\textcolor{teal}{+0.27}}$&\underline{81.67}$_{\textcolor{teal}{+0.34}}$&\underline{87.30}$_{\textcolor{teal}{+0.37}}$\\  
         \bottomrule
    \end{tabular}
    }
    \caption{Comparison between baselines and our method (in gray) across closed-source GPT-3.5 and GPT-4 models.}
    \label{table:main_results_closed_models}
\end{table*}

We also find that the majority voting based self-consistency does not always help. For example, the performance drops (25.33$\rightarrow$24.33, 24.33$\rightarrow$21.67, and 24.67$\rightarrow$23.33) on word unscrambling task. This indicates that for some tasks, due to the difficulty of problems and the inherent limitations of LLMs, majority voting based self-consistency does not always lead to improvements. Overall, our ranked voting based self-consistency generally outperforms majority voting based methods such as Few-Shot-CoT-SC and Adaptive-SC.

\noindent\textbf{7B$\sim$9B Open-Source LLMs.} The results are shown in Table~\ref{table:main_results_open_models_7b_9b}. Similarly, our methods achieved the best and second-best results, indicating that they are still effective on the medium-size LLMs of ranked voting. Across different scenarios, compared to Few-Shot-CoT-SC, LLaMa-3-8B achieves an average performance improvement of 2.68\% to 3.51\%. Compared to the 2B$\sim$4B lightweight LLMs, the overall improvement has slightly decreased, possibly due to the increase in the LLMs' inherent capacity.

\noindent\textbf{Closed-Source LLMs.} The results are shown in Table~\ref{table:main_results_closed_models}. Our method achieves an average accuracy of 76.69\% and 87.41\% on gpt-3.5-turbo-0125 and gpt-4-turbo-2024-04-09, respectively, outperforming the best baseline methods by 3.4\% and 0.48\%. These two results reflect the differences in GPT-3.5 and GPT-4. Similar to the open-source models, we see large improvements on GPT-3.5, while on the state-of-the-art GPT-4 model, the gains from voting across multiple responses decrease. Nevertheless, our ranking voting method still performs slightly better than the baselines.

\section{Analyses}
\label{section:analysis}
We further analyze our proposed method from different perspectives. Unless otherwise specified, we conduct experiments on CommonsenseQA and WikiData using LLaMa-3-8B.

\begin{figure}[t!]
  \centering
  \includegraphics[scale=0.19]{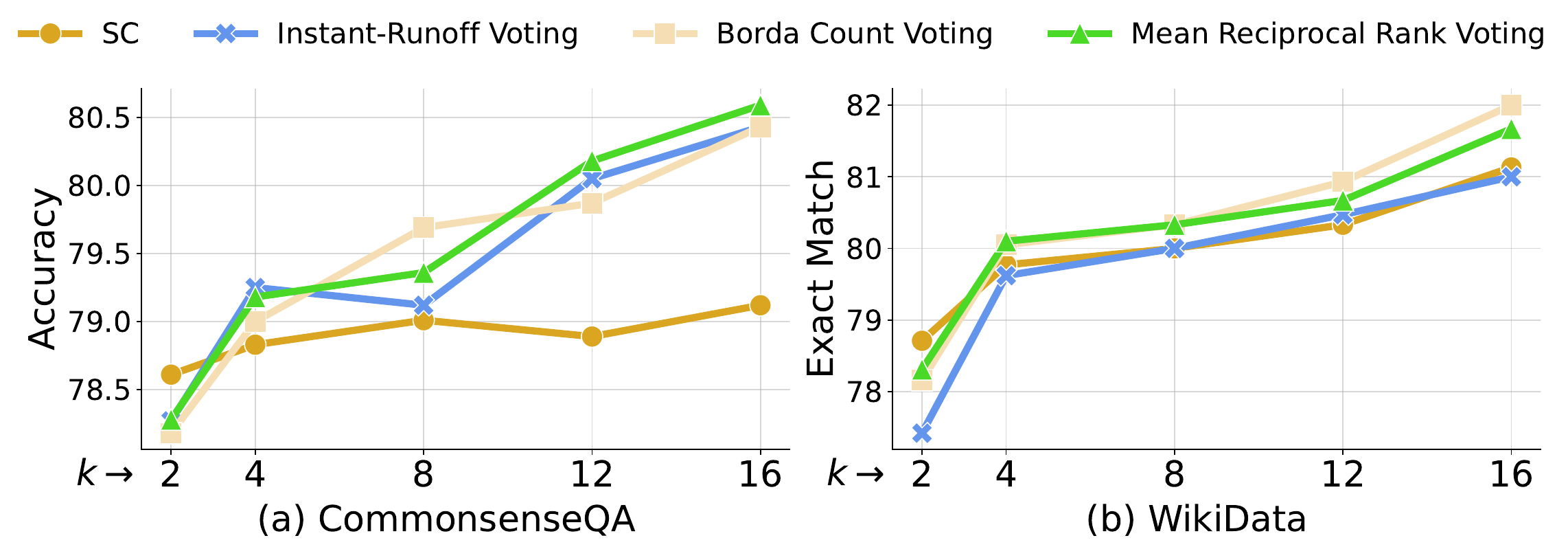}
  \caption{Comparison of methods with different number of responses $k$.}
  \label{AnalysisOfK}
\end{figure}

\noindent\textbf{The Impact of $k$.} Figure~\ref{AnalysisOfK} shows the performance of different number of samples $k$. For CommonsenseQA, under different settings of $k$, ranked voting methods consistently outperform majority voting based self-consistency. Moreover, compared with baseline with $k=12$ or more, our method achieves better results when $k=8$, and its performance improves as $k$ increases. In contrast, the baseline shows limited improvement as $k$ increases. For WikiData, the results of instant-runoff voting and baseline are similar, while the other two methods consistently achieve an accuracy improvement of 0.5$\sim$1.0 with different settings of $k$.

\begin{figure}
  \centering
  \includegraphics[scale=0.19]{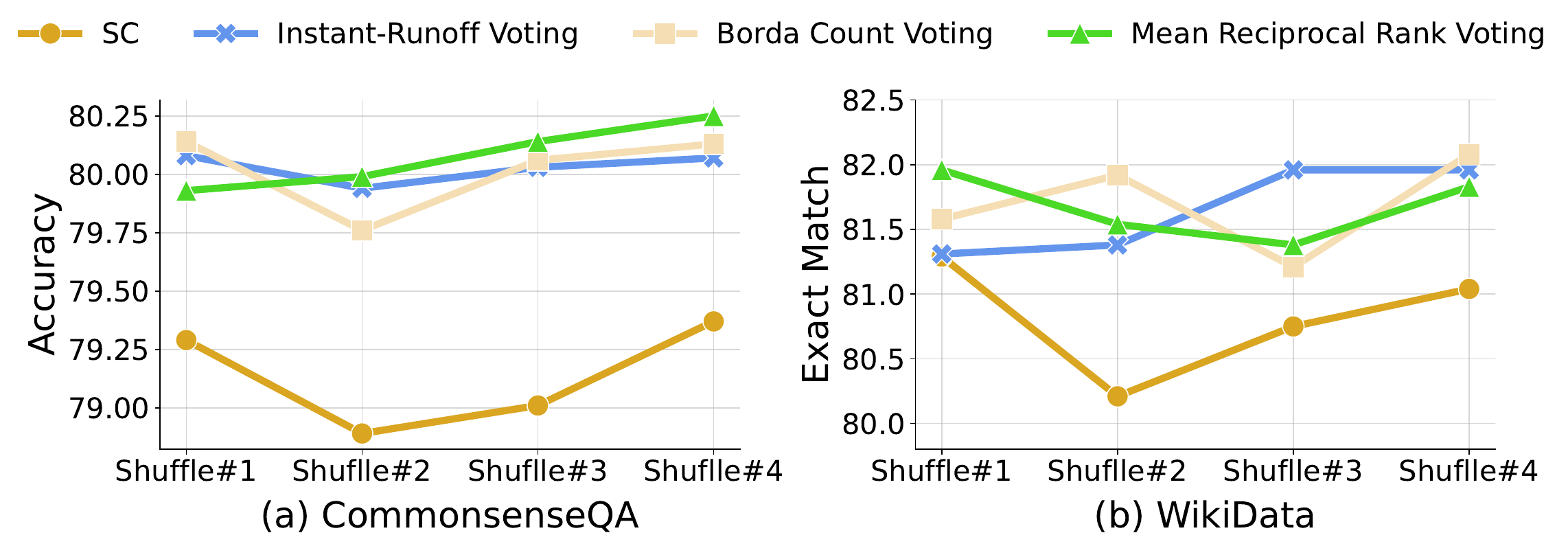}
  \caption{Comparison of methods with different shuffled examples for few-shot learning.}
  \label{AnalysisOfRobustness}
\end{figure}

\noindent\textbf{Robustness Analysis.} The performance of LLMs is relatively sensitive to the order of examples in few-shot setting, such as exhibiting some recency bias~\cite{pmlr-v139-zhao21c}. We randomly shuffled the order of few-shot ($n=8$) examples and conducted four experiments, and the results are shown in Figure~\ref{AnalysisOfRobustness}. For both CommonsenseQA and WikiData, the performance of our method surpasses that of the baseline consistently, and it exhibits relatively smaller variance, indicating that our approach leverages additional ranking information to yield more robust results.

\begin{figure}[t!]
  \centering
  \includegraphics[scale=0.59]{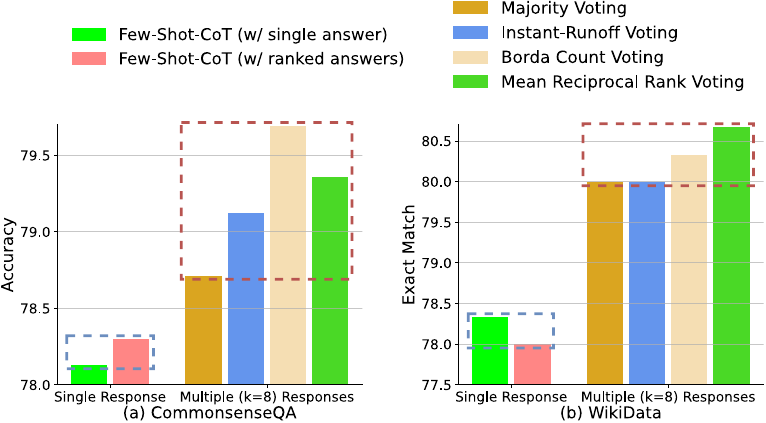}
      \caption{The impact of ranked answers (in \textcolor[rgb]{0.4235, 0.5569, 0.7490}{blue} dashed box) and ranked voting (in \textcolor[rgb]{0.7216, 0.3294, 0.3137}{red} dashed box). Ranked voting is more effective and capable of delivering substantial and reliable improvements.}
  \label{Analysis4}
\end{figure}

\noindent\textbf{Compare the Impact of Ranked Answers and the Impact of Ranked Voting.} To investigate whether the improvement stems from the ranking of answers or the ranked voting, we conducted experiments with the settings where only a single response with ranked answers is provided, taking the top-ranked answer as the final result. In this setting, the voting mechanisms are not employed. The results are shown in Figure~\ref{Analysis4}. We find that the ranked answers along do not lead to consistent improvement, even making the results worse on WikiData. The gap between ranked voting and majority voting is much larger than that between Few-Shot-CoT (w/ ranked answers) and Few-Shot-CoT (w/ single answer), indicating that ranked voting can indeed enhance the model’s performance. 

\noindent\textbf{The Impact of the Number of Candidates.}  We observed that the number of candidates can significantly affect the performance of ranked voting methods. To systematically examine this effect, we select a set of values $\mathcal{C} = {1, 2, 3, 4, 5}$ and conduct experiments using our approach under each candidate setting.

\begin{figure}[t!]
  \centering
  \includegraphics[scale=0.19]{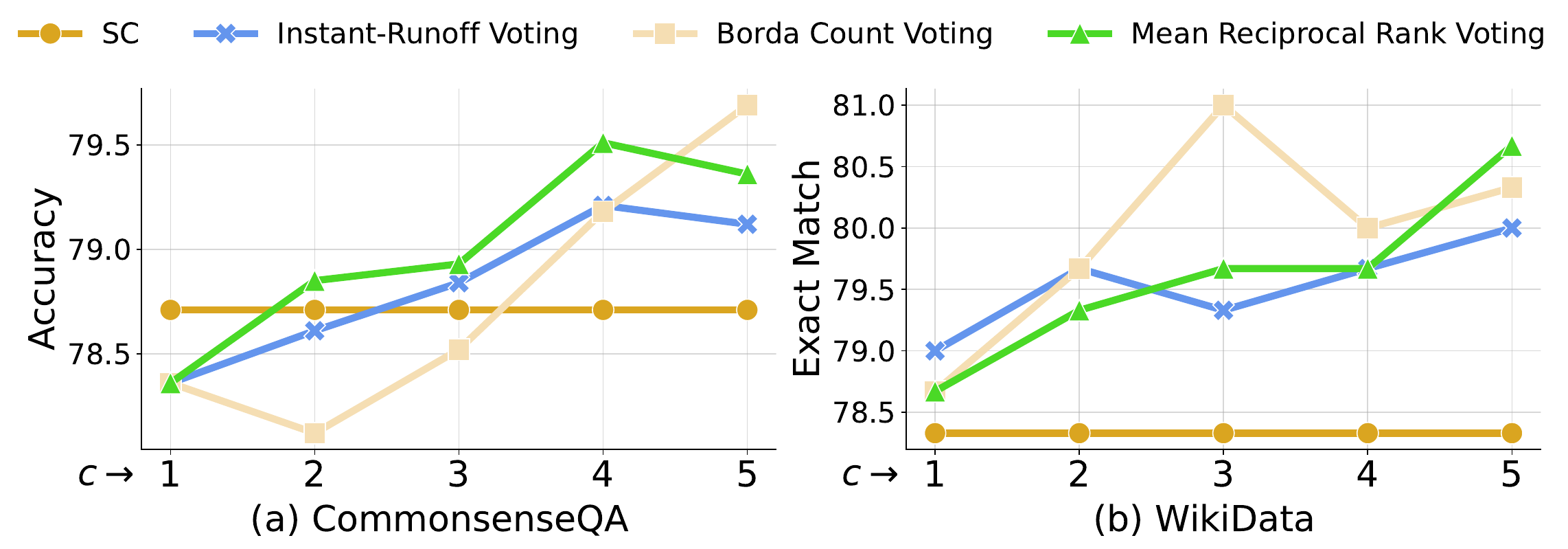}
  \caption{Performance comparison of SC and ranked voting methods (IRV, BCV, MRRV) under different candidate numbers $c$ (Since SC does not incorporate the concept of ranking candidates, its performance remains constant and thus appears as a flat line). }
  \label{AnalysisOfCandidates}
\end{figure}

As shown in Figure~\ref{AnalysisOfCandidates}, we find that the performance of our methods is positively correlated with the number of candidates $c$: as $c$ increases, the performance also improves. On CommonsenseQA, our methods surpass the baseline with only $c=2$ candidates, and by $c=4$, all ranked voting strategies consistently outperform SC. On WikiData, even with $c=1$, ranked methods already achieve better performance than SC, and their accuracy continues to rise as more candidates are considered. These results demonstrate that incorporating a richer set of candidate responses allows our ranked approaches to extract stronger reasoning signals, leading to improved robustness and generalization across tasks. Moreover, increasing the number of candidates effectively encourages the model to explore a broader solution space, promoting deeper reasoning and more diverse perspectives during the generation process.

\begin{table}[t!]
   \centering
   \scalebox{0.7}{
   \begin{tabular}{lcccc}
        \toprule
          \multirow{2.5}*{\textbf{Datasets}} & \multirow{2.5}*{\textbf{Majority Voting}}&\multicolumn{3}{c}{\textbf{Ranked Voting}} \\
          \cmidrule(lr){3-5}
          &  & \textbf{IRV} & \textbf{BCV} & \textbf{MRRV} \\
        \midrule
        CommonsenseQA & 5.08\% & 3.77\% & 4.42\% & 2.29\% \\
        WikiData & 4.33\% & 4.33\% & 4.00\% & 3.67\% \\
        \bottomrule
  \end{tabular}
  }
  \caption{The tie rates for different voting methods.}
  \label{table:TieAndRanking}
\end{table}

\noindent\textbf{The Impact of Ranking Information on Tie Situations.} Both majority voting and ranked voting may result in ties among multiple candidates, which can lead to confusing and incorrect answers. The ranked voting methods leverage ranking information, which effectively reduces the occurrence of ties, as confirmed by our experiments.

The results are shown in Table~\ref{table:TieAndRanking}, we find that there are 4.33\%$\sim$5.08\% ties in majority voting based self-consistency, while our methods reduce this to 2.29\%$\sim$4.33\%. Among the three ranked voting methods, the weighting-based methods (BCV and MRRV) are less likely to result in ties compared to the elimination-based method (IRV). The difference arises from the vote aggregation mechanisms. Weighting-based methods assign varying confidence scores to candidate answers, reducing the likelihood of ties. In contrast, IRV iteratively eliminates lower-ranked options, which can lead to ties when multiple answers have similar rankings.

\section{Related Work}

\noindent\textbf{Chain-of-Thought Reasoning.} Chain-of-thought (CoT) have shown effectiveness for eliciting the reasoning ability of LLMs through generating rationales in contexts~\citep{cot,lets-think-step-by-step,chen2023program,zhang2023cumulative}. Although the generated rationales improve the performance, previous studies focus on designing prompts for certain tasks. For example, \citet{yao2023tree} and \citet{besta2023graph} further propose tree-of-thought (ToT) and graph-of-thought (GoT) for solving search-style or elaborate problems, respectively. These works usually generate a single final answer in each trial and there is a lack of study on prompting the model to generate multiple possible answers and corresponding priority relationships.

\vspace{0.5mm}
\noindent\textbf{Self-Consistency of LLMs.} \citet{wang2023selfconsistency} propose self-consistency to improve chain-of-thought reasoning, a decoding method that improves the accuracy of LLMs by taking the majority vote among multiple answers generated through diverse reasoning paths. \citet{aggarwal-etal-2023-lets} further propose adaptive self-consistency to reduce the computational cost for obtain diverse reasoning paths, which dynamically adjusts the number of responses based on the model's confidence using a stopping criteria. \citet{xiong-etal-2023-examining} introduce inter-consistency among LLMs for multi agents collaboration. \citet{wang-etal-2024-soft} introduce soft self-consistency in long-duration interactive tasks for language model agents. \citet{huang-etal-2024-enhancing} enhance the coding ability of LLMs through multiple-perspective self-consistency. As a comparison, these methods rely on majority voting for specific tasks and ignore the possibility of multiple potential answers, which could offer useful information for obtaining the final correct answer in general reasoning and question-answering tasks.

\vspace{0.5mm}
\noindent\textbf{Ranked Voting.} Ranked voting expresses the preferences of voters by ranking multiple candidates, organizing selections on an ordinal scale, which has been widely applied in elections, competitions, and recommendation systems~\cite{myatt2007theory,colomer2013ramon}. Similar ideas have also been applied to enhance the performance of neural models. \citet{schwartz-2021-ensemble} propose using an ensemble method to aggregate predictions from multiple MRR and NDCG based models. ~\citet{liu2024aligning} proposes pairwise-preference search to address bias in text evaluation by constructing a global ranking. \citet{zhao-etal-2024-electoral} introduces general electoral decision-making interface to enhance collaboration among multiple LLM agents through ordinal preferential voting. In recent years, ranking systems have also been valued and proposed to be introduced into more NLP benchmarks~\cite{NEURIPS2022_ac4920f4,rofin-etal-2023-votenrank} and Chatbot arenas~\cite{min2025improving}. \citet{tang-etal-2024-found} similarly leverage multiple generations via permutation self-consistency to enhance ranking robustness, though their focus is on mitigating positional bias rather than vote-based aggregation.

\section{Conclusion}
We introduce ranked voting based self-consistency to enhance the reasoning performance of large language models. Our method outperforms traditional majority voting techniques by generating and incorporating ranking information among multiple candidate answers during the reasoning process. The proposed method improves the reliability of self-consistency and boosts the performance. Extensive experiments demonstrate the effectiveness of our method, as it consistently outperforms baselines across a variety of multiple-choice and open-ended question-answering datasets, using both open-source and closed-source LLMs.

\section*{Limitations}
First, ranked voting has a rich history and comes in many forms. In our main experiments, we demonstrate three effective ranked voting strategies for large language models, however, there are some other ranked voting methods show limited improvement in datasets we used. For example, in the Appendix~\ref{appendix:moreresults}, we introduce approval voting and find that this method performs similarly to majority voting when applied to ranked answers from models. Second, as shown in Figure~\ref{AnalysisOfCandidates}, the number of candidates in the ranked answers may have an impact on the performance. Increasing the number of candidates allows a broader range of potential options to be considered, which could improve the accuracy of the final decision. In this work, we enable the model to autonomously rank a small set of options or open-ended generated candidates (usually 4$\sim$5), without imposing additional constraints by designing prompts to expand the candidate pool.

\bibliography{custom}

\appendix

\section{Examples of Datasets}
\label{appendix:examples}

We list examples of different datasets in Table~\ref{table:instance}. For CommonsenseQA, we evaluated all instances from the validation set since the test set does not provide answers. For ARC-Challenge, AQUA-RAT, and Date Understanding, we tested on all instances from the test set. For WikiData and Word Unscrambling, we randomly sampled 300 instances from the test set for evaluation.

\begin{table}[t!]
	\centering
 \small
	\begin{tabular}{p{7.2cm}}
	    \toprule
            \textbf{CommonsenseQA} (1221)\\
            Question: What must someone do before they shop?\\
            Options: (A) get money (B) have money (C) bring cash (D) go to market (E) bring cash\\
            Answer: (A).\\
            \midrule
            \textbf{ARC-Challenge} (1172)\\
            Question: There is a very slight change in gravity at the top of a mountain. Which property would most likely be less at the top of a mountain? \\
            Options: (A) mass (B) weight (C) body density (D) body temperature\\
            Answer: (B).\\
            \midrule
            \textbf{AQUA-RAT} (254)\\
            Question: I have a money pouch containing Rs. 700. There are equal number of 25 paise coins, 50 paise coins and one rupee coins. How many of each are there?\\
            Options: (A) 453 (B) 651 (C) 400 (D) 487 (E) 286\\
            Answer: (C).\\
            \midrule
            \textbf{WikiData} (300)\\
            Question: The language of The Rise and Fall of Ziggy Stardust and the Spiders from Mars is \\
            Answer: English.\\
            \midrule	
            \textbf{Date Understanding} (369) \\
            Question: Jane got her job in 2016. Today is her 3-year work anniversary. She still remember that on Dec 2, her second day at work, she spilled coffee on her laptop. What is the date today in MM/DD/YYYY?\\
            Answer: 12/01/2019.\\
            \midrule	
            \textbf{Word Unscrambling} (300) \\
            Question: The word eptnerrreenu is a scrambled version of the English word\\
            Answer: entrepreneur.\\          
            
        \bottomrule
	\end{tabular}
	\caption{Examples of each dataset, the numbers in brackets denote the number of evaluation questions.}
	\label{table:instance}
\end{table}

\section{Checkpoints of Models}
\label{appendix:checkpoints}
The checkpoints of open-source models in our experiments are shown in Table~\ref{checkpoints}.

\begin{table*}[t!]
\centering
    \resizebox{0.9\linewidth}{!}{
    \begin{tabular}{ll}
    \toprule 
    \textbf{Model}&\textbf{Checkpoints}\\
         \midrule
         LLaMA-3.2-3B&\url{https://huggingface.co/meta-llama/Llama-3.2-3B-Instruct}\\
         LLaMA-3-8B&\url{https://huggingface.co/meta-llama/Meta-Llama-3-8B-Instruct}\\
         Qwen-2.5-3B&\url{https://huggingface.co/Qwen/Qwen2.5-3B-Instruct}\\
         Qwen-2.5-7B&\url{https://huggingface.co/Qwen/Qwen2.5-7B-Instruct}\\
         Gemma-2-2B&\url{https://huggingface.co/google/gemma-2-2b-it}\\
         Gemma-2-9B&\url{https://huggingface.co/google/gemma-2-9b-it}\\
         Phi-3-4B&\url{https://huggingface.co/microsoft/Phi-3-mini-4k-instruct}\\
         Phi-3-7B&\url{https://huggingface.co/microsoft/Phi-3-small-8k-instruct}\\
        \bottomrule
    \end{tabular}}
    \caption{Checkpoints of open-source models in our experiments. Note that LLaMA-3 series only include 8B and 70B models, while LLaMA-3.2 series include 1B, 3B, 11B, and 90B models. Therefore, in our experiments, we use LLaMA-3-8B and LLaMA-3.2-3B for validating on the LLaMA models.}
    \label{checkpoints}
\end{table*}

\section{More Results of Other Ranked Voting based Methods}
\label{appendix:moreresults}
\noindent\textbf{Approval Voting.} In addition to the three voting methods discussed in the main pages, we introduce an additional approach called approval voting \cite{laslier2003approval}. This voting system enables participants to express their stance on an issue—whether to approve or disapprove. To streamline this process, approval voting assumes that higher-ranked answers are more likely to be accepted as correct. Therefore, we establish a threshold $t$: answers ranked before this threshold are considered approved, while those ranked after are deemed disapproved:
\begin{equation}
\begin{array}{l}
\begin{aligned}
    &\mathcal{A}^{r}_{1:t} = \mathcal{A}^{r_1}, \mathcal{A}^{r_2}, \ldots, \mathcal{A}^{r_{t}},\\
    &\mathcal{A}^{r}_\text{approved} = \mathcal{A}^{r}_{1:t}.
\end{aligned}
\end{array}
\end{equation}

Next, we conduct majority voting to select the answer with the most supporters, where each approval vote carries equal weight in the majority voting process. In the experiments, we set the threshold $t = 2$.

\noindent\textbf{Results.} Table~\ref{table:moreresults} shows the performance of approval voting. It is evident that in certain cases, approval voting outperforms self-consistency and can even achieve the best performance. For example, in the WikiData task, LLaMA-3-8B attains the highest performance using approval voting. However, this method has its limitations as well, since it is not particularly stable. Although it performs well in some situations, it may sometimes underperform compared to Few-Shot-CoT. 

\begin{table*}[t!]
  \centering
  \scalebox{0.62}{
  \begin{tabular}{lccccccc}
        \toprule
          \multirow{2.5}*{\textbf{Method}}&\multicolumn{3}{c}{\textbf{Multiple-Choice QA } (Accuracy)}&\multicolumn{3}{c}{\textbf{Open-Ended QA} (Exact Match)}&\multirow{2.5}*{\textbf{Average}}\\
          \cmidrule(lr){2-4}\cmidrule(lr){5-7}
          &\textbf{CommonsenseQA}&\textbf{ARC-Challenge}&\textbf{AQUA-RAT}&\textbf{WikiData}&\textbf{Date Understanding}&\textbf{Word Unscrambling}&\\
        \midrule
        \multicolumn{8}{c}{(\textsc{LLaMA-3-8B})}\\
  Few-Shot-CoT&\underline{78.13}\phantom{$_{+0.00}$}&\underline{84.13}\phantom{$_{+0.00}$}&62.99\phantom{$_{+0.00}$}&78.33\phantom{$_{+0.00}$}&\underline{70.19}\phantom{$_{+0.00}$}&\underline{25.67}\phantom{$_{+00.00}$}&66.67\phantom{$_{+0.00}$}\\
   Few-Shot-CoT-SC&\textbf{78.71}\phantom{$_{+0.00}$}&\textbf{86.77}\phantom{$_{+0.00}$}&\textbf{66.93}\phantom{$_{+0.00}$}&\underline{80.00}\phantom{$_{+0.00}$}&\textbf{71.82}\phantom{$_{+0.00}$}&24.00\phantom{$_{+00.00}$}&\underline{68.04}\phantom{$_{+0.00}$}\\
   \rowcolor{gray!20}
   Approval Voting&77.48$_{\textcolor{red}{-1.23}}$&83.96$_{\textcolor{red}{-2.81}}$&\underline{65.35}$_{\textcolor{red}{-1.58}}$&\textbf{80.67}$_{\textcolor{teal}{+0.67}}$&69.38$_{\textcolor{red}{-2.44}}$&\textbf{34.67}$_{\textcolor{teal}{+10.67}}$&\textbf{68.58}$_{\textcolor{teal}{+0.54}}$\\
         \midrule
         \multicolumn{8}{c}{(\textsc{Qwen-2.5-7B})}\\
  Few-Shot-CoT&83.29\phantom{$_{+0.00}$}&\underline{87.80}\phantom{$_{+0.00}$}&\underline{80.31}\phantom{$_{+0.00}$}&68.00\phantom{$_{+0.00}$}&68.29\phantom{$_{+0.00}$}&17.00\phantom{$_{+0.00}$}&67.45\phantom{$_{+0.00}$}\\
  Few-Shot-CoT-SC&\textbf{84.28}\phantom{$_{+0.00}$}&\textbf{89.16}\phantom{$_{+0.00}$}&\textbf{84.65}\phantom{$_{+0.00}$}&\underline{70.33}\phantom{$_{+0.00}$}&\textbf{70.73}\phantom{$_{+0.00}$}&\underline{21.00}\phantom{$_{+0.00}$}&\textbf{70.03}\phantom{$_{+0.00}$}\\
   \rowcolor{gray!20}
   Approval Voting&\underline{83.46}$_{\textcolor{red}{-0.82}}$&87.20$_{\textcolor{red}{-1.96}}$&76.77$_{\textcolor{red}{-7.88}}$&\textbf{78.33}$_{\textcolor{teal}{+8.00}}$&\underline{69.65}$_{\textcolor{red}{-1.08}}$&\textbf{22.33}$_{\textcolor{teal}{+1.33}}$&\underline{69.62}$_{\textcolor{red}{-0.41}}$\\
         \midrule
         \multicolumn{8}{c}{(\textsc{Gemma-2-9B})}\\
  Few-Shot-CoT&\underline{80.75}\phantom{$_{+0.00}$}&88.05\phantom{$_{+0.00}$}&\underline{63.39}\phantom{$_{+0.00}$}&78.00\phantom{$_{+0.00}$}&\underline{77.24}\phantom{$_{+0.00}$}&\underline{45.67}\phantom{$_{+0.00}$}&72.18\phantom{$_{+0.00}$}\\
  Few-Shot-CoT-SC&80.10\phantom{$_{+0.00}$}&\textbf{89.59}\phantom{$_{+0.00}$}&\textbf{67.32}\phantom{$_{+0.00}$}&\underline{81.33}\phantom{$_{+0.00}$}&\textbf{78.05}\phantom{$_{+0.00}$}&44.33\phantom{$_{+0.00}$}&\underline{73.45}\phantom{$_{+0.00}$}\\
   \rowcolor{gray!20}
   Approval Voting&\textbf{82.64}$_{\textcolor{teal}{+2.54}}$&\underline{88.99}$_{\textcolor{red}{-0.60}}$&\textbf{67.32}$_{\textcolor{teal}{+0.00}}$&\textbf{82.00}$_{\textcolor{teal}{+0.67}}$&76.15$_{\textcolor{red}{-1.90}}$&\textbf{46.33}$_{\textcolor{teal}{+2.00}}$&\textbf{73.91}$_{\textcolor{teal}{+0.46}}$\\
         \midrule
         \multicolumn{8}{c}{(\textsc{Phi-3-7B})}\\
    Few-Shot-CoT&80.43\phantom{$_{+0.00}$}&\underline{91.47}\phantom{$_{+0.00}$}&65.75\phantom{$_{+0.00}$}&\underline{79.33}\phantom{$_{+0.00}$}&68.56\phantom{$_{+0.00}$}&\underline{27.67}\phantom{$_{+0.00}$}&68.87\phantom{$_{+0.00}$}\\
    Few-Shot-CoT-SC&\textbf{81.24}\phantom{$_{+0.00}$}&\textbf{91.89}\phantom{$_{+0.00}$}&\textbf{73.62}\phantom{$_{+0.00}$}&\underline{79.33}\phantom{$_{+0.00}$}&\textbf{73.71}\phantom{$_{+0.00}$}&\textbf{28.33}\phantom{$_{+0.00}$}&\textbf{71.35}\phantom{$_{+0.00}$}\\
   \rowcolor{gray!20}
   Approval Voting&\underline{80.92}$_{\textcolor{red}{-0.32}}$&90.36$_{\textcolor{red}{-1.53}}$&\underline{67.72}$_{\textcolor{red}{-5.90}}$&\textbf{80.67}$_{\textcolor{teal}{+1.34}}$&\underline{70.73}$_{\textcolor{red}{-2.98}}$&\underline{27.67}$_{\textcolor{red}{-0.66}}$&\underline{69.68}$_{\textcolor{red}{-1.67}}$\\ 
       \bottomrule
  \end{tabular}
  }
  \caption{Results for approval voting using open-source LLMs, where the overall improvement is limited.}
  \label{table:moreresults}
\end{table*}

\section{Results of Multiple Experimental Runs}
\label{appendix:multiple experiments}
The standard deviation results for multiple experimental runs are shown in Table~\ref{table:results_with_standard_deviation}. The comparable and relatively small standard deviations between baseline and ours indicate the stability of the implementation and our ranked voting methods.
\begin{table*}[t!]
    \centering
    \scalebox{0.67}{
    \begin{tabular}{lcccccc}
            \toprule
                \multirow{3}*{\textbf{Method}}&\multicolumn{3}{c}{\textbf{Multiple-Choice QA}}&\multicolumn{3}{c}{\textbf{Open-Ended QA}}\\
                \cmidrule(lr){2-4}\cmidrule(lr){5-7}
                &\textbf{CommonsenseQA}&\textbf{ARC-Challenge}&\textbf{AQUA-RAT}&\textbf{WikiData}&\textbf{Date Understanding}&\textbf{Word Unscrambling}\\
            \midrule
            \multicolumn{7}{c}{(\textsc{LLaMA-3.2-3B})}\\
        Few-Shot-CoT-SC & 73.46$_{ \pm 0.3 }$ & 80.54$_{\pm 0.4}$ & 61.81$_{\pm 0.3}$ & 73.67$_{\pm 0.2}$ & 60.98$_{\pm 0.5}$ & 24.33$_{\pm 0.6}$\\
        \rowcolor{gray!20}
        Instant-Runoff Voting & \textbf{74.86}$_{ \pm 0.2 }$ & \underline{81.31}$_{\pm 0.2}$ & \underline{69.29}$_{\pm 0.3}$ & 74.00$_{\pm 0.4}$ & \textbf{64.23}$_{\pm 0.4}$ & 27.67$_{\pm 0.7}$\\
        \rowcolor{gray!20}
        Borda Count Voting & \underline{74.69}$_{ \pm 0.2 }$ & 80.55$_{\pm 0.2}$ & 67.32$_{\pm 0.2}$ & \textbf{77.00}$_{\pm 0.4}$ & 60.16$_{\pm 0.8}$ & \underline{28.33}$_{\pm 0.7}$\\
        \rowcolor{gray!20}
        Mean Reciprocal Rank Voting & 74.45$_{ \pm 0.1 }$ & \textbf{81.40}$_{\pm 0.3}$ & \textbf{71.26}$_{\pm 0.3}$ & \underline{76.00}$_{\pm 0.4}$ & 62.60$_{\pm 0.9}$ & \textbf{29.00}$_{\pm 0.6}$\\  
            \midrule
            \multicolumn{7}{c}{(\textsc{Qwen-2.5-3B})}\\
        Few-Shot-CoT-SC & 77.89$_{ \pm 0.2 }$ & 76.96$_{\pm 0.18}$ & 77.95$_{\pm 0.2}$ & 72.67$_{\pm 0.2}$ & 47.97$_{\pm 0.4}$ & 7.33$_{\pm 0.02}$\\
        \rowcolor{gray!20}
        Instant-Runoff Voting & \textbf{78.95}$_{ \pm 0.2 }$ & \underline{83.45}$_{\pm 0.3}$ & 79.13$_{\pm 0.1}$ & 74.67$_{\pm 0.1}$ & \textbf{60.70}$_{\pm 0.5}$ & \underline{11.67}$_{\pm 0.5}$\\
        \rowcolor{gray!20}
        Borda Count Voting & 78.38$_{ \pm 0.2 }$ & 83.19$_{\pm 0.5}$ & \textbf{80.31}$_{\pm 0.4}$ & \textbf{76.33}$_{\pm 0.1}$ & 58.81$_{\pm 0.5}$ & 11.33$_{\pm 0.6}$\\
        \rowcolor{gray!20}
        Mean Reciprocal Rank Voting & \underline{78.79}$_{ \pm 0.3 }$ & \textbf{83.70}$_{\pm 0.4}$ & \underline{79.92}$_{\pm 0.1}$ & \underline{75.33}$_{\pm 0.1}$ & \underline{60.43}$_{\pm 0.4}$ & \textbf{12.33}$_{\pm 0.6}$\\  
            \midrule
            \multicolumn{7}{c}{(\textsc{Gemma-2-2B})}\\
        Few-Shot-CoT-SC & 69.29$_{ \pm 0.2 }$ & 71.42$_{\pm 0.2}$ & 36.22$_{\pm 0.2}$ & 73.67$_{\pm 0.2}$ & 37.67$_{\pm 0.9}$ & 21.67$_{\pm 0.5}$\\
        \rowcolor{gray!20}
        Instant-Runoff Voting & \textbf{71.50}$_{ \pm 0.2 }$ & \textbf{74.74}$_{\pm 0.4}$ & \textbf{44.49}$_{\pm 0.1}$ & 75.67$_{\pm 0.4}$ & \underline{37.94}$_{\pm 0.2}$ & \textbf{23.00}$_{\pm 0.3}$\\
        \rowcolor{gray!20}
        Borda Count Voting & 70.93$_{ \pm 0.2 }$ & \underline{73.12}$_{\pm 0.3}$ & 42.52$_{\pm 0.2}$ & \textbf{77.67}$_{\pm 0.5}$ & 37.40$_{\pm 0.4}$ & 22.00$_{\pm 0.5}$\\
        \rowcolor{gray!20}
        Mean Reciprocal Rank Voting & \underline{71.42}$_{ \pm 0.2 }$ & 72.18$_{\pm 0.3}$ & \underline{42.91}$_{\pm 0.1}$ & \underline{77.33}$_{\pm 0.5}$ & \textbf{39.02}$_{\pm 0.4}$ & \underline{22.67}$_{\pm 0.4}$\\
            \midrule
            \multicolumn{7}{c}{(\textsc{Phi-3-4B})}\\
        Few-Shot-CoT-SC & 75.84$_{ \pm 0.3 }$ & 90.13$_{\pm 0.2}$ & 73.62$_{\pm 0.5}$ & 77.33$_{\pm 0.2}$ & \underline{66.12}$_{\pm 0.5}$ & \underline{23.33}$_{\pm 0.5}$\\
        \rowcolor{gray!20}
        Instant-Runoff Voting & 78.54$_{ \pm 0.3 }$ & \textbf{90.70}$_{\pm 0.3}$ & \underline{74.41}$_{\pm 0.1}$ & \underline{79.00}$_{\pm 0.4}$ & \textbf{66.40}$_{\pm 0.4}$ & \textbf{27.00}$_{\pm 1.1}$\\
        \rowcolor{gray!20}
        Borda Count Voting & \underline{78.71}$_{ \pm 0.3 }$ & 88.23$_{\pm 0.2}$ & \textbf{75.20}$_{\pm 0.3}$ & 78.67$_{\pm 0.3}$ & 63.96$_{\pm 0.9}$ & \textbf{27.00}$_{\pm 1.3}$\\
        \rowcolor{gray!20}
        Mean Reciprocal Rank Voting & \textbf{78.95}$_{ \pm 0.3 }$ & \underline{90.44}$_{\pm 0.2}$ & \textbf{75.20}$_{\pm 0.3}$ & \textbf{80.00}$_{\pm 0.3}$ & 65.58$_{\pm 0.4}$ & \textbf{27.00}$_{\pm 1.2}$\\
            \midrule
            \multicolumn{7}{c}{(\textsc{LLaMA-3-8B})}\\
        Few-Shot-CoT-SC & 78.71$_{ \pm 0.2 }$ & 86.77$_{\pm 0.2}$ & 66.93$_{\pm 0.3}$ & 80.00$_{\pm 0.3}$ & \underline{71.82}$_{\pm 0.8}$ & 24.00$_{\pm 0.6}$\\
        \rowcolor{gray!20}
        Instant-Runoff Voting & 79.12$_{ \pm 0.2 }$ & \underline{87.29}$_{\pm 0.2}$ & \underline{74.02}$_{\pm 0.3}$ & 80.00$_{\pm 0.2}$ & \textbf{72.09}$_{\pm 0.3}$ & \underline{35.00}$_{\pm 0.6}$\\
        \rowcolor{gray!20}
        Borda Count Voting & \textbf{79.69}$_{ \pm 0.2 }$ & 87.12$_{\pm 0.2}$ & 70.87$_{\pm 0.3}$ & \underline{80.33}$_{\pm 0.4}$ & 71.00$_{\pm 0.4}$ & \textbf{35.33}$_{\pm 1.0}$\\
        \rowcolor{gray!20}
        Mean Reciprocal Rank Voting & \underline{79.36}$_{ \pm 0.2 }$ & \textbf{87.46}$_{\pm 0.2}$ & \textbf{75.20}$_{\pm 0.3}$ & \textbf{80.67}$_{\pm 0.4}$ & 71.27$_{\pm 0.6}$ & \textbf{35.33}$_{\pm 1.1}$\\  
            \midrule
            \multicolumn{7}{c}{(\textsc{Qwen-2.5-7B})}\\
        Few-Shot-CoT-SC & 84.28$_{ \pm 0.1 }$ & \underline{89.16}$_{\pm 0.2}$ & 84.65$_{\pm 0.1}$ & 70.33$_{\pm 0.5}$ & \underline{70.73}$_{\pm 0.7}$ & 21.00$_{\pm 0.3}$\\
        \rowcolor{gray!20}
        Instant-Runoff Voting & \underline{84.77}$_{ \pm 0.2 }$ & \underline{89.16}$_{\pm 0.2}$ & \underline{85.04}$_{\pm 0.1}$ & 77.67$_{\pm 0.3}$ & \textbf{71.27}$_{\pm 0.3}$ & \textbf{24.00}$_{\pm 0.4}$\\
        \rowcolor{gray!20}
        Borda Count Voting & 84.52$_{ \pm 0.3 }$ & \textbf{89.25}$_{\pm 0.2}$ & 84.25$_{\pm 0.3}$ & \textbf{79.33}$_{\pm 0.3}$ & \underline{70.73}$_{\pm 0.4}$ & \underline{23.33}$_{\pm 0.3}$\\
        \rowcolor{gray!20}
        Mean Reciprocal Rank Voting & \textbf{84.93}$_{ \pm 0.2 }$ & 87.80$_{\pm 0.2}$ & \textbf{85.83}$_{\pm 0.2}$ & \underline{78.67}$_{\pm 0.3}$ & \textbf{71.27}$_{\pm 0.4}$ & \textbf{24.00}$_{\pm 0.3}$\\  
            \midrule
            \multicolumn{7}{c}{(\textsc{Gemma-2-9B})}\\
        Few-Shot-CoT-SC & 80.10$_{ \pm 0.1 }$ & \underline{89.59}$_{\pm 0.2}$ & 67.32$_{\pm 0.2}$ & 81.33$_{\pm 0.3}$ & 78.05$_{\pm 0.2}$ & 44.33$_{\pm 0.4}$\\
        \rowcolor{gray!20}
        Instant-Runoff Voting & \textbf{83.21}$_{ \pm 0.1 }$ & \textbf{89.68}$_{\pm 0.2}$ & \textbf{72.83}$_{\pm 0.2}$ & 82.67$_{\pm 0.4}$ & \underline{78.59}$_{\pm 0.2}$ & \textbf{46.33}$_{\pm 0.4}$\\
        \rowcolor{gray!20}
        Borda Count Voting & 82.96$_{ \pm 0.1 }$ & 89.33$_{\pm 0.1}$ & 70.87$_{\pm 0.2}$ & \textbf{83.33}$_{\pm 0.1}$ & \textbf{79.40}$_{\pm 0.1}$ & \underline{46.00}$_{\pm 0.4}$\\
        \rowcolor{gray!20}
        Mean Reciprocal Rank Voting & \underline{83.05}$_{ \pm 0.2 }$ & 89.25$_{\pm 0.1}$ & \underline{72.44}$_{\pm 0.2}$ & \underline{83.00}$_{\pm 0.2}$ & 78.32$_{\pm 0.1}$ & \textbf{46.33}$_{\pm 0.4}$\\  
            \midrule
            \multicolumn{7}{c}{(\textsc{Phi-3-7B})}\\
        Few-Shot-CoT-SC & 81.24$_{ \pm 0.4 }$ & 91.89$_{\pm 0.1}$ & 73.62$_{\pm 0.3}$ & 79.33$_{\pm 0.2}$ & 73.71$_{\pm 0.5}$ & 28.33$_{\pm 0.8}$\\
        \rowcolor{gray!20}
        Instant-Runoff Voting & 81.98$_{ \pm 0.2 }$ & \textbf{92.24}$_{\pm 0.1}$ & \textbf{77.17}$_{\pm 0.2}$ & \underline{80.00}$_{\pm 0.4}$ & \underline{75.34}$_{\pm 0.4}$ & \textbf{30.33}$_{\pm 0.8}$\\
        \rowcolor{gray!20}
        Borda Count Voting & \textbf{82.31}$_{ \pm 0.2 }$ & 91.81$_{\pm 0.2}$ & \underline{75.20}$_{\pm 0.3}$ & \underline{80.00}$_{\pm 0.5}$ & \textbf{75.61}$_{\pm 0.5}$ & 28.33$_{\pm 0.7}$\\
        \rowcolor{gray!20}
        Mean Reciprocal Rank Voting & \underline{82.15}$_{ \pm 0.2 }$ & \underline{91.98}$_{\pm 0.1}$ & \textbf{77.17}$_{\pm 0.2}$ & \textbf{81.00}$_{\pm 0.4}$ & \underline{75.34}$_{\pm 0.3}$ & \underline{28.67}$_{\pm 0.7}$\\   
            \midrule
            \multicolumn{7}{c}{(\textsc{gpt-3.5-turbo-0125})}\\
        Few-Shot-CoT-SC & 79.61$_{ \pm 0.2 }$ & 87.54$_{\pm 0.1}$ & 69.49$_{\pm 0.3}$ & 79.00$_{\pm 0.1}$ & 57.99$_{\pm 0.4}$ & 54.50$_{\pm 0.2}$\\
        \rowcolor{gray!20}
        Instant-Runoff Voting & \underline{80.84}$_{ \pm 0.1 }$ & \underline{89.33}$_{\pm 0.1}$ & \underline{70.67}$_{\pm 0.2}$ & \underline{82.00}$_{\pm 0.3}$ & \textbf{68.97}$_{\pm 0.5}$ & \underline{66.50}$_{\pm 0.3}$\\
        \rowcolor{gray!20}
        Borda Count Voting & 80.92$_{ \pm 0.1 }$ & \textbf{89.59}$_{\pm 0.1}$ & 68.90$_{\pm 0.2}$ & 81.50$_{\pm 0.2}$ & 68.70$_{\pm 0.5}$ & \textbf{67.67}$_{\pm 0.4}$\\
        \rowcolor{gray!20}
        Mean Reciprocal Rank Voting & \textbf{81.00}$_{ \pm 0.1 }$ & 89.25$_{\pm 0.1}$ & \textbf{71.06}$_{\pm 0.2}$ & \textbf{82.33}$_{\pm 0.3}$ & \underline{68.83}$_{\pm 0.5}$ & \textbf{67.67}$_{\pm 0.3}$\\  
            \midrule
            \multicolumn{7}{c}{(\textsc{gpt-4-turbo-2024-04-09})}\\
        Few-Shot-CoT-SC & \underline{87.39}$_{ \pm 0.1 }$ & 95.30$_{\pm 0.1}$ & \underline{86.22}$_{\pm 0.2}$ & \underline{83.00}$_{\pm 0.1}$ & \underline{88.35}$_{\pm 0.3}$ & 81.33$_{\pm 0.0}$\\
        \rowcolor{gray!20}
        Instant-Runoff Voting & \textbf{87.47}$_{ \pm 0.1 }$ & \textbf{97.01}$_{\pm 0.1}$ & \textbf{86.61}$_{\pm 0.1}$ & \underline{83.00}$_{\pm 0.2}$ & \underline{88.35}$_{\pm 0.2}$ & \textbf{82.00}$_{\pm 0.0}$\\
        \rowcolor{gray!20}
        Borda Count Voting & 87.31$_{ \pm 0.1 }$ & \underline{96.84}$_{\pm 0.1}$ & 85.04$_{\pm 0.2}$ & \textbf{83.33}$_{\pm 0.2}$ & \underline{88.35}$_{\pm 0.2}$ & \underline{81.67}$_{\pm 0.1}$\\
        \rowcolor{gray!20}
        Mean Reciprocal Rank Voting & 87.31$_{ \pm 0.1 }$ & \textbf{97.01}$_{\pm 0.1}$ & \underline{86.22}$_{\pm 0.1}$ & \underline{83.00}$_{\pm 0.2}$ & \textbf{88.62}$_{\pm 0.2}$ & \underline{81.67}$_{\pm 0.1}$\\  
            \bottomrule
    \end{tabular}
    }
    \caption{The standard deviation results in subscripts of multiple experimental runs.}
    \label{table:results_with_standard_deviation}
\end{table*}

\section{Case Study}
\label{appendix:case}
We show cases of the ranked voting on CommonsenseQA in Table~\ref{case1}, Table~\ref{case2}, and Table~\ref{case3}.

\begin{table*}[t!]
    \centering
     \scalebox{0.7}{
    \begin{tabular}{p{20cm}}
    \toprule
    \rowcolor[gray]{0.95} 
    \textbf{CommonsenseQA} \\
    \midrule
    \textbf{Question:} Dance can be elegant and specific, or you can just have fun and what? \\
    \textbf{Options:} (A) falling down (B) trip (C) fall down (D) move around (E) celebrate. \\
    \textbf{Ground-Truth Label:} D \\
        \midrule
    \textbf{Self-Consistency Answer through Majority Voting}: A. \textcolor{red}{\xmark}\\
    \midrule
    \textbf{Our Ranked Answers:}\\
    \textbf{Response 1:} The answer should focus on an action that represents a casual, carefree approach, contrasting with the elegance and precision of dance. ``Trip'' fits well because it conveys a playful or carefree moment that can happen when you're not focusing on being precise. This aligns with the idea of just having fun. The ranking of options by likelihood is: B > D > E > C > A. \\

    \textbf{Response 2:} The answer should be the action that is the opposite of being elegant and specific, which implies a more casual or carefree approach. Of the above choices, the closest match is ``just have fun and fall down'', which implies a more playful and carefree attitude. The ranking of options by likelihood is: A > D > E > C > B. \\  
    
    \textbf{Response 3:} The answer should be the phrase that describes the opposite of elegance, which is a more casual and carefree way of dancing. Of the above choices, ``falling down'' implies a loss of control and a more playful approach to dancing. The ranking of options by likelihood is: A > E > D > C > B. \\  
    
    \textbf{Response 4:} The answer should be a phrase that describes an action that is carefree and fun, which is a common aspect of dancing. The ranking of options by likelihood is: E > D > A > C > B. \\  
    
    \textbf{Response 5:} The answer should be the action that is often associated with having fun while dancing. Of the above choices, ``move around'' is a general action that is often done while having fun and dancing, without any specific technique or elegance. The ranking of options by likelihood is: D > E > A > C > B. \\  
    
    \textbf{Response 6:} The answer should be the opposite of ``elegant and specific'', which implies a more carefree and casual attitude towards dancing. Of the above choices, ``just have fun'' is closely related to ``move around'' in a playful way. The ranking of options by likelihood is: D > E > A > C > B. \\  
    
    \textbf{Response 7:} The answer should be the action that is often associated with having fun while dancing. Of the above choices, ``have fun'' is often associated with ``celebrate''. The ranking of options by likelihood is: E > D > A > C > B. \\  
    
    \textbf{Response 8:} The answer should be the opposite of ``elegant'' and ``specific'', which implies a more casual or playful approach to dance. The ranking of options by likelihood is: A > E > D > C > B. \\
    \midrule
    \textbf{Instant-Runoff Voting}: D. \textcolor{teal}{\cmark}\\
    \midrule
     \textbf{Instant-Runoff Voting Procedure:}\\
    \textbf{Round 1:} \\
    
    B > D > E > C > A $\times$ 1\\
    
    A > D > E > C > B $\times$ 1\\
    
    A > E > D > C > B $\times$ 2\\
    
    E > D > A > C > B $\times$ 2\\
    
    D > E > A > C > B $\times$ 2\\
    
    Count: A (3 times), B (1 times), C (0 times), D (2 times), E (2 times). No option gets more than half the votes, and C has the fewest, so C is eliminated.\\
    \textbf{Round 2:} \\
    B > D > E > A $\times$ 1\\

    A > D > E > B $\times$ 1\\
    
    A > E > D > B $\times$ 2\\
    
    E > D > A > B $\times$ 2\\
    
    D > E > A > B $\times$ 2\\
    
    Count: A (3 times), B (1 times), D (2 times), E (2 times). No option gets more than half the votes, and B has the fewest, so B is eliminated.\\
    \textbf{Round 3:} \\
    D > E > A $\times$ 1\quad$\xrightarrow{merge}$\quad D > E > A $\times$ 3\\
    
    A > D > E $\times$ 1\quad\phantom{$\xrightarrow{merge}$}\quad A > D > E $\times$ 1\\
    
    A > E > D $\times$ 2\quad\phantom{$\xrightarrow{merge}$}\quad A > E > D $\times$ 2\\
    
    E > D > A $\times$ 2\quad\phantom{$\xrightarrow{merge}$}\quad E > D > A $\times$ 2\\
    
    D > E > A $\times$ 2\quad\phantom{$\xrightarrow{merge}$}\\
    
    Count: A (3 times), D (3 times), E (2 times). No option gets more than half the votes, and E has the fewest, so E is eliminated. \\
    \textbf{Round 4:} \\
    D > A $\times$ 1\quad$\xrightarrow{merge}$\quad D > A $\times$ 5\\
    
    A > D $\times$ 1\quad\phantom{$\xrightarrow{merge}$}\quad A > D $\times$ 3\\
    
    A > D $\times$ 2\quad\phantom{$\xrightarrow{merge}$}\\
    
    D > A $\times$ 2\quad\phantom{$\xrightarrow{merge}$}\\
    
    D > A $\times$ 2\quad\phantom{$\xrightarrow{merge}$}\\
    
    Count: A (3 times), D (5 times). D gets more than half the votes, so D is the winner. \\
    \bottomrule
    \end{tabular}}
    \caption{Example of instant-runoff voting on CommonsenseQA.}
    \label{case1}
\end{table*}

\begin{table*}[t!]
    \centering
     \scalebox{0.7}{
    \begin{tabular}{p{20cm}}
    \toprule
    \rowcolor[gray]{0.95} 
    \textbf{CommonsenseQA} \\
    \midrule
    \textbf{Question:} Danny found an old film in a sealed what? \\
    \textbf{Options:} (A) clingfilm (B) disneyland (C) cave (D) cabinet (E) movie. \\
    
    \textbf{Ground-Truth Label:} D \\
        \midrule
    \textbf{Self-Consistency Answer through Majority Voting}: A. \textcolor{red}{\xmark}\\
    \midrule
    \textbf{Our Ranked Answers:}\\
    \textbf{Response 1:} The answer should be the container where the film was found. Of the above choices, clingfilm and cabinet are both plausible containers, but sealed clingfilm is more likely to preserve an old film. The ranking of options by likelihood is: A > D > C > E > B. \\

    \textbf{Response 2:} The answer should be the place where an old film would likely be stored. Of the above choices, a sealed cabinet makes the most sense for storing old films. The ranking of options by likelihood is: D > C > A > E > B. \\  
    
    \textbf{Response 3:} The answer should be a place where an old film could be found, likely in a sealed container. Of the above choices, a cabinet is a common place to store things, and it could be sealed. The ranking of options by likelihood is: D > C > A > E > B. \\  
    
    \textbf{Response 4:} The answer should be the item that is used to seal something. Of the above choices, clingfilm is used to seal things. The ranking of options by likelihood is: A > D > C > E > B. \\  
    
    \textbf{Response 5:} The answer should be a container or packaging that can be sealed. Of the above choices, clingfilm is a type of plastic wrap that can be sealed. The ranking of options by likelihood is: A > D > C > E > B. \\  
    
    \textbf{Response 6:} The answer should be the container that is typically sealed to keep things inside. Of the above choices, clingfilm is a type of plastic wrap used to seal things. The ranking of options by likelihood is: A > D > C > E > B. \\  
    
    \textbf{Response 7:} The answer should be a container or object that can be sealed. Of the above choices, clingfilm (also known as plastic wrap) is commonly used to seal items. The ranking of options by likelihood is: A > D > C > E > B. \\  
    
    \textbf{Response 8:} The answer should be the container where an old film could be stored and sealed. Of the above choices, the most likely container is a cabinet. The ranking of options by likelihood is: D > C > A > E > B. \\

    \midrule
    \textbf{Borda Count Voting}: D. \textcolor{teal}{\cmark}\\
    \midrule
    \textbf{Borda Count Voting Calculation:} \\
    A: 5 + 3 + 3 + 5 + 5 + 5 + 5 + 3 = 34\\
    \vspace{1pt}
    B: 1 + 1 + 1 + 1 + 1 + 1 + 1 + 1 = 8\\
    \vspace{1pt}
    C: 3 + 4 + 4 + 3 + 3 + 3 + 3 + 4 = 27\\
    \vspace{1pt}
    D: 4 + 5 + 5 + 4 + 4 + 4 + 4 + 5 = 35 \textcolor{teal}{\cmark}\\
    \vspace{1pt}
    E: 2 + 2 + 2 + 2 + 2 + 2 + 2 + 2 = 16\\
    \bottomrule
    \end{tabular}}
    \caption{Example of Borda count voting voting on CommonsenseQA.}
    \label{case2}
\end{table*}

\begin{table*}[t!]
    \centering
     \scalebox{0.7}{
    \begin{tabular}{p{20cm}}
    \toprule
    \rowcolor[gray]{0.95} 
    \textbf{CommonsenseQA} \\
    \midrule
    \textbf{Question:} The man acted ridiculous at the funeral, what attitude should he have taken? \\
    \textbf{Options:} (A) straightforward (B) serious (C) solemn (D) somber (E) funny. \\
    \textbf{Ground-Truth Label:} C \\
        \midrule
    \textbf{Self-Consistency Answer through Majority Voting}: B. \textcolor{red}{\xmark}\\
    \midrule
    \textbf{Our Ranked Answers:}\\
    \textbf{Response 1:} The most likely attitude to take at a funeral is solemn. This word implies a sense of respect and seriousness that is appropriate for such a setting. The other options, such as straightforward, serious, and somber, are all related to seriousness and respect, but ``funny'' would be inappropriate and insensitive. The ranking of options by likelihood is: C > D > A > E > B. \\

    \textbf{Response 2:} The most appropriate attitude for the man to take at a funeral would be ``serious'' or ``somber''. Funeral are solemn occasions requiring a display of respect for the deceased and the grieving process. Funny or straight-forward would be inappropriate. The ranking of options by likelihood is: B > C > D > E > A. \\  
    
    \textbf{Response 3:} The most appropriate demeanor at a funeral is one of solemnity, as it is a moment for grieving and honoring the departed. Words like straightforward, serious, and somber all reflect an attitude of respect and mourning, while funny would be seen as unsuitable and disrespectful. The ranking of options by likelihood is: C > D > E > A > B. \\  
    
    \textbf{Response 4:} The most likely answer is (B) serious, as a funeral is a somber occasion and a serious attitude is appropriate and respectful. The other options, like straightforward, solemn, and somber, are less likely to be the correct response to a funeral. The ranking of options by likelihood is: B > C > A > D > E. \\  
    
    \textbf{Response 5:} The most likely attitude to take at a funeral is (B) serious, as this is a time of mourning and respect, and a funeral is a solemn occasion. The other options, such as straightforward, somber, and funny, are less appropriate for such an event. The ranking of options by likelihood is: B > E > A > C > D. \\  
    
    \textbf{Response 6:} The most appropriate attitude for the man to have taken at a funeral would be ``somber'', as it reflects a sense of respect and sadness appropriate for such a solemn occasion. The other options, while potentially applicable in other contexts, are not the best match for a funeral, which is generally a time for mourning and reflection. The ranking of options by likelihood is: D > C > E > A > B. \\  
    
    \textbf{Response 7:} The most likely attitude to take at a funeral is solemn, as it is a time for mourning and respect for the deceased. Straightforward, serious, and somber are all synonyms for showing respect and mourning, while funny would be inappropriate and disrespectful. The ranking of options by likelihood is: C > B > D > A > E. \\  
    
    \textbf{Response 8:} The most likely attitude the man should have taken is (B) serious, as funerals are somber occasions requiring respect and solemnity. The other options, such as straightforward, solemn, and somber, are less appropriate for such an event. The ranking of options by likelihood is: B > C > D > E > A. \\
    \midrule
    \textbf{Mean Reciprocal Rank Voting}: C. \textcolor{teal}{\cmark}\\
    \midrule
    \textbf{Mean Reciprocal Rank Voting Calculation:}\\
    A: $\frac{1}{8}(\frac{1}{3} + \frac{1}{5} + \frac{1}{4} + \frac{1}{3} + \frac{1}{3} + \frac{1}{4} + \frac{1}{4} + \frac{1}{5}) = 0.26875$\\
    \vspace{1pt}
    B: $\frac{1}{8}(\frac{1}{5} + 1 + \frac{1}{5} + 1 + 1 + \frac{1}{5} + \frac{1}{2} + 1) = 0.6375$\\
    \vspace{1pt}
    C: $\frac{1}{8}(1 + \frac{1}{2} + 1 + \frac{1}{2} + \frac{1}{4} + \frac{1}{2} + 1 + \frac{1}{2}) = 0.65625$ \textcolor{teal}{\cmark}\\
    \vspace{1pt}
    D: $\frac{1}{8}(\frac{1}{2} + \frac{1}{3} + \frac{1}{2} + \frac{1}{4} + \frac{1}{5} + 1 + \frac{1}{3} + \frac{1}{3}) = 0.43125$\\
    \vspace{1pt}
    E: $\frac{1}{8}(\frac{1}{4} + \frac{1}{4} + \frac{1}{3} + \frac{1}{5} + \frac{1}{2} + \frac{1}{3} + \frac{1}{5} + \frac{1}{4}) \approx 0.28958$\\
    \bottomrule
    \end{tabular}}
    \caption{Example of mean reciprocal rank voting on CommonsenseQA.}
    \label{case3}
\end{table*}

\end{document}